\documentclass[10pt,twocolumn,letterpaper]{article}

\usepackage{iccv}
\usepackage{times}
\usepackage{epsfig}
\usepackage{graphicx}
\usepackage{amsmath}
\usepackage{amssymb}
\usepackage{microtype}
\usepackage{comment}
\usepackage{float}
\usepackage[title]{appendix}

\usepackage[pagebackref=true,breaklinks=true,letterpaper=true,colorlinks,bookmarks=false]{hyperref}

\iccvfinalcopy 

\newcommand{\boldstartspace}[1]{\vspace{0.1in}\noindent\textbf{#1.}}

\ificcvfinal\fi
\begin{document}

\title{Inverse Rendering for Complex Indoor Scenes:\\ Shape, Spatially-Varying Lighting and SVBRDF from a Single Image}

\author{
Zhengqin Li{$^\dagger$} \hspace{-0.15cm},
\hspace{0.05cm} 
Mohammad Shafiei{$^\dagger$} \hspace{-0.15cm},
\hspace{0.05cm} 
Ravi Ramamoorthi{$^\dagger$} \hspace{-0.15cm},
\hspace{0.05cm} 
Kalyan Sunkavalli{$^\ddagger$} \hspace{-0.15cm},
\hspace{0.05cm} 
Manmohan Chandraker{$^\dagger$}
\\[0.3cm]
{$^\dagger$}University of California, San Diego {\hspace{1cm}} {$^\ddagger$}Adobe Research
}

\maketitle

\begin{abstract}
We propose a deep inverse rendering framework for indoor scenes. From a single RGB image of an arbitrary indoor scene, we create a complete scene reconstruction, estimating shape, spatially-varying lighting, and spatially-varying, non-Lambertian surface reflectance. To train this network, we augment the SUNCG indoor scene dataset with real-world materials and render them with a fast, high-quality, physically-based GPU renderer to create a large-scale, photorealistic indoor dataset. Our inverse rendering network incorporates physical insights -- including a spatially-varying spherical Gaussian lighting representation, a differentiable rendering layer to model scene appearance, a cascade structure to iteratively refine the predictions and a bilateral solver for refinement -- allowing us to jointly reason about shape, lighting, and reflectance. Experiments show that our framework outperforms previous methods for estimating individual scene components, which also enables various novel applications for augmented reality, such as photorealistic object insertion and material editing. Code and data will be made publicly available.   
\end{abstract}

\vspace{-0.2cm}
\section{Introduction}
\label{sec::introduction}
\vspace{-0.2cm}

A long-standing problem in computer vision is to reconstruct a scene---including its shape, lighting, and material properties---from a single image. This is an ill-posed task: these scene factors interact in complex ways to form images and multiple combinations of these factors may produce the same image~\cite{adelson96perception}. As a result, previous work has often focused on subsets of this problem---shape reconstruction, illumination estimation, intrinsic images, etc.---or on restricted settings---single objects or objects from a specific class.

\begin{figure}
\centering
\includegraphics[width=3.2in]{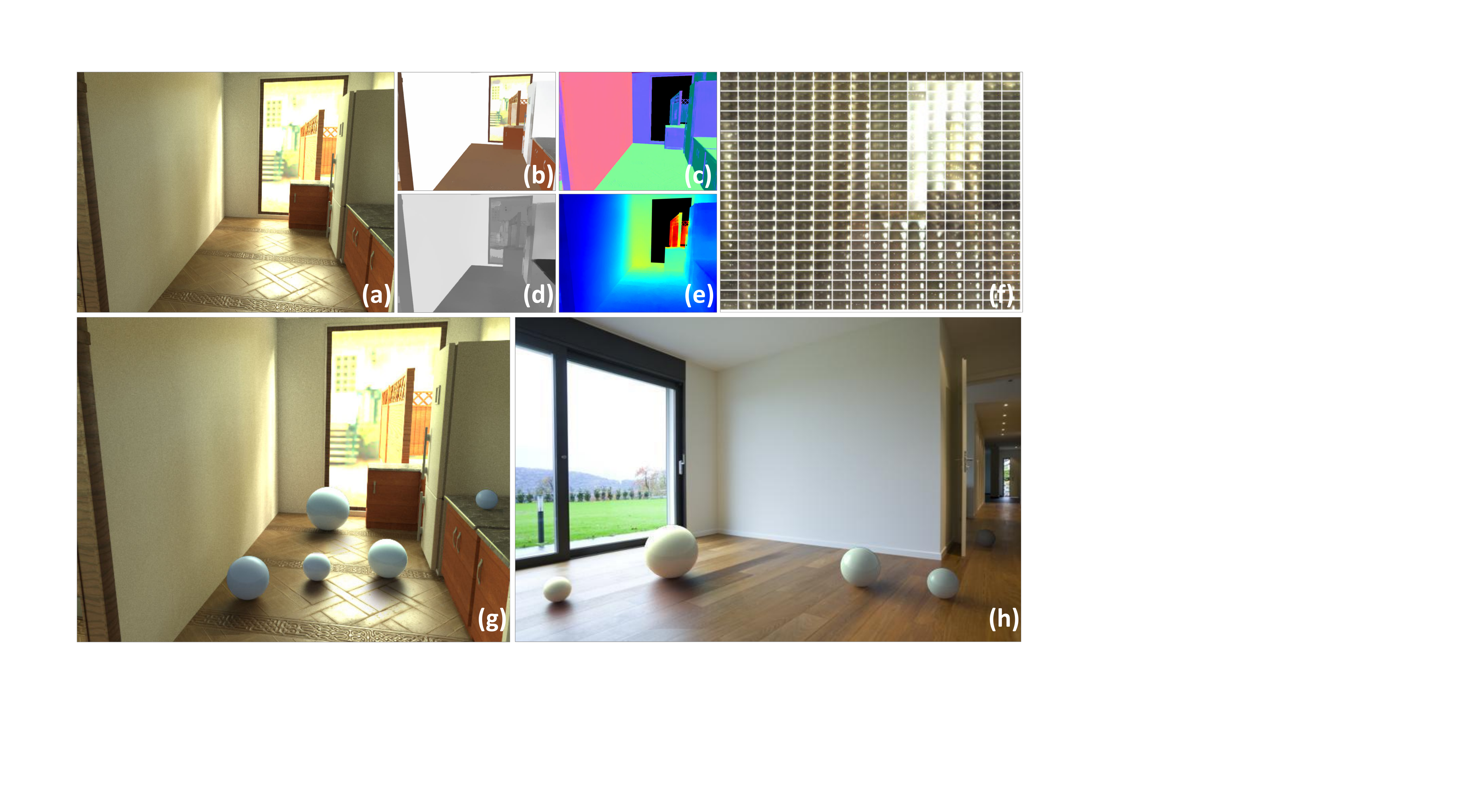}
\caption{Given a single image of an indoor scene (a), we recover its diffuse albedo (b), normals (c), specular roughness (d), depth (e) and spatially-varying lighting (f). We train our model with high-quality synthetic images with photorealistic materials. By incorporating physical insights into our network structure, our predictions are of high enough equality to support applications like object insertion, even for specular objects (g) and in real images (h). Note the completely shadowed sphere on the extreme right.}
\label{fig:teaser}
\vspace{-0.3cm}
\end{figure}

\begin{figure*}
\centering
\includegraphics[width=1.0\linewidth]{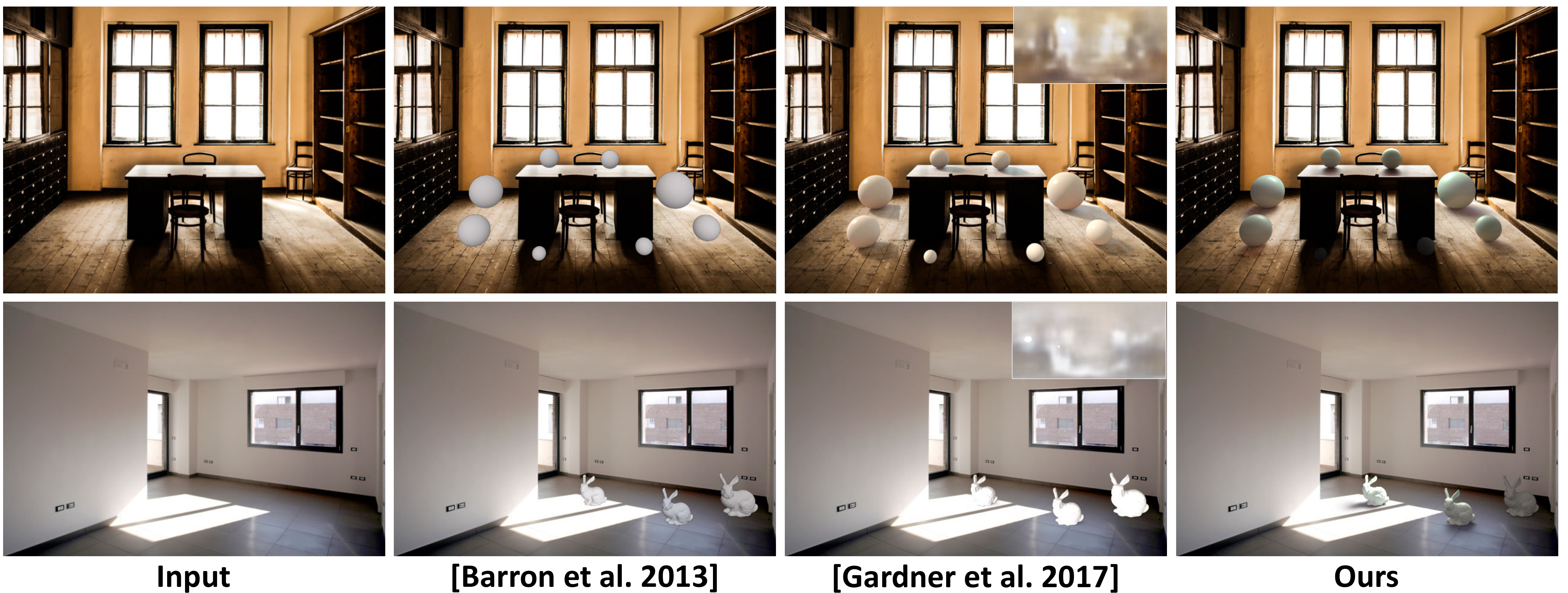}
\caption{Comparison of object insertion results on a real image. Barron et al.~\cite{barron2013intrinsic} predict spatially varying $\log$ shading from a RGBD image, but their lighting representation does not preserve high frequency signal and cannot be used to render shadows and inter-reflections. Gardner et al.~\cite{gardner2017indoor} predict a single lighting for the whole scene and therefore cannot model spatially varying indoor lighting.
In contrast, our method solves the indoor scene inverse rendering problem in a holistic way, which results in photorealistic object insertion. The quality of our output may be visualized in a video for the example at the top, generated without any temporal constraints, at this \href{https://drive.google.com/file/d/1qD3xhK-NQuNu3ZbWeYE5annryZ9DCF3k/view?usp=sharing}{link}.}
\label{fig:early_demo_objInsert}
\vspace{-0.2cm}
\end{figure*}

\begin{figure}
\centering
\includegraphics[width=3.2in]{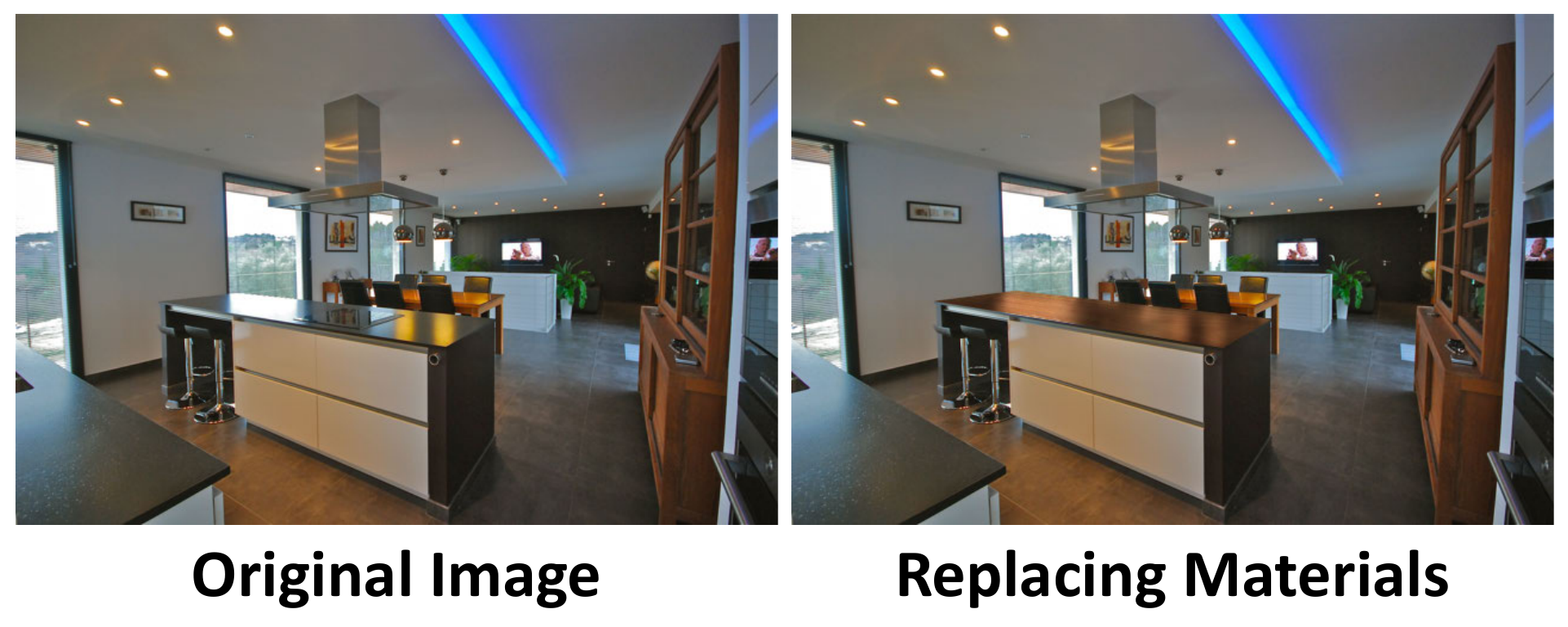}
\caption{A material editing example where we replace a material (on the surface of the kitchen counter-top) with a different one. Note the specular highlights on the surface, which can not be handled by conventional intrinsic decomposition methods since they do not recover the lighting direction. In contrast, we recover spatially-varying lighting and material properties.}
\label{fig:early_demo_matEdit}
\vspace{-0.2cm}
\end{figure}

Our goal is a solution to a more general problem: from a single RGB image of an arbitrary indoor scene captured under uncontrolled conditions, we seek to reconstruct geometry, spatially-varying surface reflectance, and spatially-varying lighting. This is a challenging setting: indoor scenes demonstrate the entire range of real-world appearance, including arbitrary geometry and layouts, localized light sources that lead to complex spatially-varying lighting effects, and complex, non-Lambertian surface reflectance. In this work we take a step towards providing a completely automatic, robust, holistic solution to this problem, thereby enabling a range of scene understanding and editing tasks. For example, in Figure~\ref{fig:teaser}(h), we use our scene reconstruction to enable photorealistic virtual object insertion; note how the inserted glossy spheres have realistic shading, shadowing caused by scene occlusions, and even reflections from the scene. 

Driven by the success of deep learning methods on similar scene inference tasks (geometric reconstruction~\cite{eigen2015depth}, lighting estimation~\cite{gardner2017indoor}, material recognition~\cite{bell15minc}), we propose training a deep convolutional neural network to regress these scene parameters from an input image. Ideally, the trained network should learn meaningful priors on these scene factors, and jointly model the interactions between them. In this work, we present two major contributions to address this. 

Training deep neural networks requires large-scale, labeled training data. While datasets of real world geometry exist~\cite{dai2017scannet,chang17matterport3D}, capturing real world lighting and surface reflectance at scale is non-trivial. Therefore, we use the SUNCG synthetic indoor scene dataset~\cite{song2016ssc} that contains a large, diverse set of indoor scenes with complex geometry. However, the materials used in SUNCG are not realistic and the rendered images~\cite{zhang2016physically} are noisy. We address this by replacing SUNCG materials with high-quality, photorealistic SVBRDFs from a high-quality 3D material dataset \cite{adobestockdata}. We automatically map these SVBRDFs to SUNCG materials using deep features from a material estimation network, thus preserving scene semantics. We render the new scenes using a GPU-based global illumination renderer, to create high-quality input images. We also render the new scene reflectance and lighting and use it (along with the original geometry) to supervise our inverse rendering network. 

An inverse rendering network would have to learn a model of image formation. The forward image formation model is well understood, and has been used in simple settings like planar scenes and single objects~\cite{deschaintre2018single,li2018learning,li2018materials,liu2017material}. Indoor scenes are more complicated and exhibit challenging light transport effects like occlusions and inter-reflections. We address this by using a local lighting model---spatially-varying spherical gaussians (SVSGs). This bakes light transport effects directly into the lighting and makes rendering a purely local computation. We leverage this to design a fast, differentiable, \emph{in-network} rendering layer that takes our geometry, SVBRDFs and SVSGs and computes radiance values. During training, we render our predictions and backpropagate the error through the rendering layer; this fixes the forward model, allowing the network to focus on the inverse task.  

To the best of our knowledge, our work is the first demonstration of scene-level inverse rendering that truly accounts for complex geometry, lighting, materials, and light transport. Moreover, we demonstrate that we achieve results on par with state-of-the-art methods focused on specific tasks. For example, the diffuse albedo reconstructed using our method is competitive with a state-of-the-art intrinsic image method. Most importantly, by truly decomposing a scene into physically-based scene factors, we enable novel capabilities like photorealistic 3D object insertion and scene editing in images acquired in-the-wild. Figure \ref{fig:early_demo_objInsert} shows two object insertion examples on real indoor scene images. Since our method solves the inverse rendering problem in a holistic way, it achieves superior performances on object insertion compared with previous state-of-the-art methods \cite{gardner2017indoor,barron2013intrinsic}. Figure \ref{fig:early_demo_matEdit} shows a material editing example, where we replace the material of a planar surface in a real image. Note that our method preserves spatially-varying specular highlights after changing the material. Such visual effects cannot be handled by traditional intrinsic decomposition methods.

\section{Related Work}
\label{sec::relatedWorks}

The problem of reconstructing shape, reflectance, and illumination from images has a long history in vision. It has been studied under different forms, such as intrinsic images (reflectance and shading from an image)~\cite{barrow1978intrinsic} and shape-from-shading (shape, and sometimes reflectance, from an image)~\cite{horn1989sfs}. 
Here, we focus on \emph{single} image methods.

\boldstartspace{Single objects}
Many inverse rendering methods focus on reconstructing single objects. Even this problem is ill-posed and many methods assume some knowledge of the object in terms of known lighting~\cite{oxholm2012shape,johnson2011natural} or geometry~\cite{lombardi2012reflectance,romeiro2010}. Other methods focus on specific object classes; for example, there are many methods that reconstruct facial shape, reflectance, and illumination using low-dimensional face models~\cite{blanz93morphable}. Recent methods have leveraged deep networks to reconstruct complex SVBRDFs from single images (captured under unknown environments) of simpler planar scenes~\cite{deschaintre2018single,li2018materials}, objects of a specific class~\cite{liu2017material} or homogeneous BRDFs~\cite{meka2018lime}. Other methods address illumination estimation~\cite{georgoulis2017illumination}. We tackle the much harder case of large-scale scene modeling and do not assume scene information.

Barron and Malik~\cite{barron13sirfs} propose an optimization-based approach with hand-crafted priors to reconstruct shape, Lambertian reflectance, and distant illumination from a single image of an arbitrary object. Li et al.~\cite{li2018learning} tackle the same problem with a deep network and an object-specific rendering layer. Extending these methods to scenes is non-trivial because the light transport is significantly more complex.

\boldstartspace{Large-scale scenes}
Previous work has looked at recognizing materials in indoor scenes~\cite{bell15minc} and decomposing indoor images into reflectance and shading layers~\cite{bell2014intrinsic,li2018cgintrinsics}. Techniques have also been proposed for single image geometric reconstruction~\cite{eigen2015depth} and lighting estimation~\cite{hold-geoffroy2017outdoor,gardner2017indoor}. These methods estimate only one scene factor without modeling the rest of scene appearance, as we do. 

Barron and Malik~\cite{barron2013intrinsic} reconstruct Lambertian reflectance and spatially-varying lighting but require an RGBD input image. Karsch et al.~\cite{karsch-siga-11} propose a full-fledged scene reconstruction method that estimates geometry, Lambertian reflectance, and 3D lighting from a single image; however, they rely on extensive user input to annotate geometry and initialize lighting. Subsequently, they propose an automatic, rendering-based optimization method~\cite{karsch-tog-14} that estimates all these scene factors. However, they rely on strong heuristics for their method that are often violated in the real world leading to errors in their estimates. In contrast, we propose a deep network that learns to predict geometry, complex SVBRDFs, and lighting in an end-to-end fashion.

\boldstartspace{Datasets}
The success of deep networks has led to an interest in datasets for supervised training. This includes real world scans~\cite{dai2017scannet,chang17matterport3D} and synthetic shape~\cite{chang2015shapenet} and scene~\cite{song2016ssc} datasets. All these datasets are either missing or have unrealistic material and lighting specifications. We build on the SUNCG dataset to improve its quality in this regard.

\boldstartspace{Differentiable rendering}
A number of recent deep inverse rendering methods have incorporated in-network, differentiable rendering layers that are customized for simple settings: faces~\cite{shu2017neuralface,tewari2017mofa}, planar surfaces~\cite{deschaintre2018single,li2018materials}, single objects~\cite{liu2017material,li2018learning}. Some recent work has proposed differentiable general-purpose global illumination renderers~\cite{li2018diffMC,che2018inverse}; unlike our more specialized, fast rendering layer, these are too expensive to use for neural network training.

\begin{figure*}[!!t]
\begin{center}
\begin{minipage}[t]{0.55\linewidth}
\raisebox{-4.3cm}
{
\includegraphics[width=4.2in]{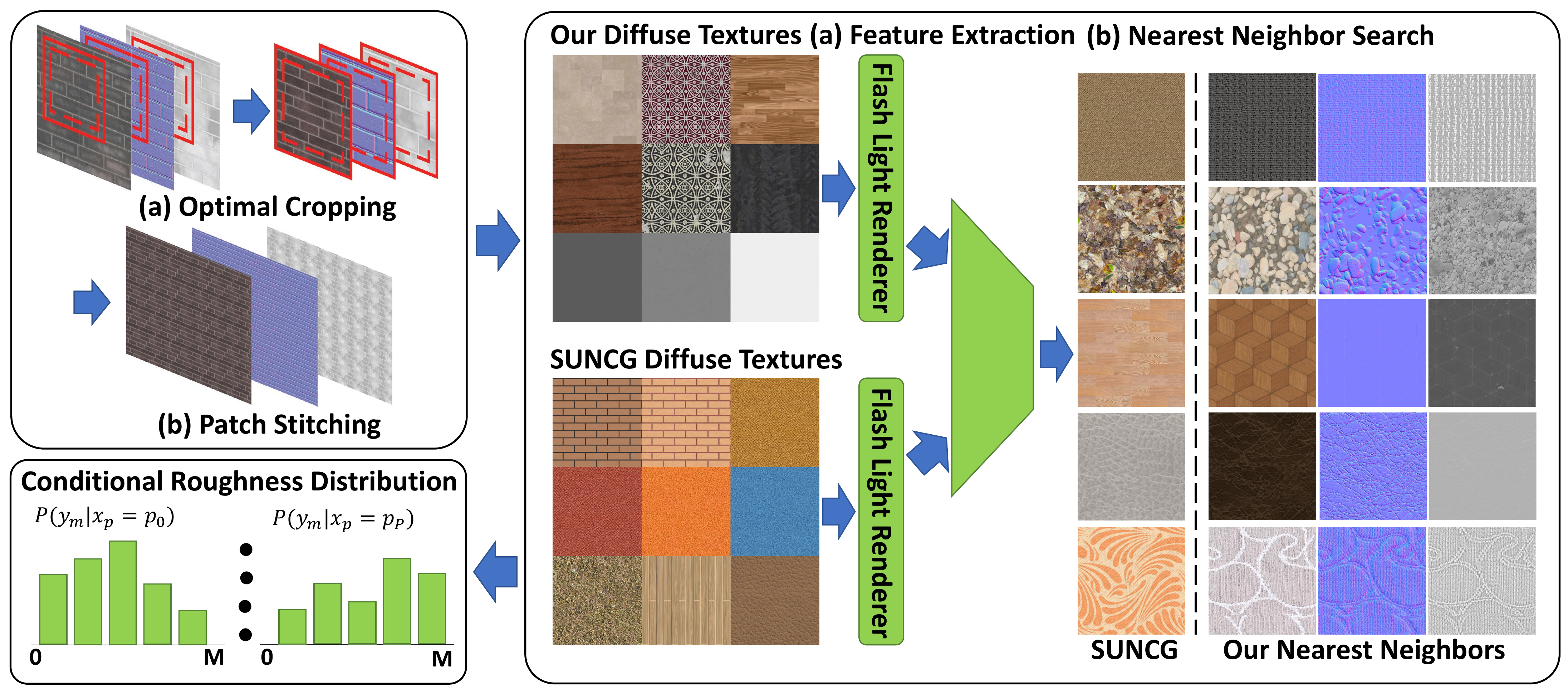}
}
\end{minipage}\hfill
\begin{minipage}[t]{0.36\linewidth}
\caption{\small
The pipeline of material mapping from original SUNCG materials with Phong BRDF to our microfacet BRDF. It has three steps. (Top left) Tileable texture synthesis to turns our SVBRDF textures into tileable ones. (Right) Spatially varying material mapping from SUNCG dataset with diffuse texture to our materials. (Bottom left) Homogeneous material mapping to convert specular parameters of homogeneous materials in SUNCG from Phong BRDF to our microfacet BRDF.
\label{fig:mat_mapping}
}
\end{minipage}
\end{center}
\end{figure*}

\section{Dataset for Complex Indoor Scenes}
\label{sec::dataset}

A large-scale dataset is crucial for solving the complex task of inverse rendering in indoor scenes. It is extremely difficult, if at all possible, to acquire large-scale ground truth with spatially-varying material, lighting and global illumination. Thus, we resort to rendering a synthetic dataset, but must overcome significant challenges to ensure utility for handling real indoor scenes at test time. Existing datasets for indoor scenes are rendered with simpler assumptions on material and lighting. In this section, we describe our approach to photorealistically map our microfacet materials to SUNCG geometries \cite{song2017semantic}, while preserving semantics. Further, rendering images with SVBRDF and global illumination, as well as ground truth for spatially-varying lighting, is computationally intensive, for which we design a custom GPU-accelerated renderer that outspeeds Mitsuba on a modern 16-core CPU by an order of magnitude.

\subsection{Mapping photorealistic materials to SUNCG}
Our goal is to map our materials to SUNCG geometries in a semantically meaningful way. The original materials in SUNCG dataset are represented by a Phong BRDF model \cite{phong1975illumination} which is not suitable for complex materials \cite{ngan2005experimental}. Our materials, on the other hand, are represented by a physically motivated microfacet BRDF model \cite{karis2013real}, which consists of 1332 materials with high resolution 4096 $\times $4096 SVBRDF textures \footnote{Please refer to Appendix \ref{sec:BRDFModel} for details}. This mapping problem is non-trivial: (i) specular lobes in SUNCG are not realistic \cite{ngan2005experimental,sun2018connect}, (ii) an optimization-based fitting collapses due to local minima leading to serious over-fitting when used for learning and (iii) we must replace materials with similar semantic types while being consistent with geometry, for example, replace material on walls with other paints and on sofas with other fabrics. Thus, we devise a three-stage pipeline, summarized in Figure \ref{fig:mat_mapping}.

\vspace{-0.4cm}
\paragraph{Step 1: Tileable texture synthesis}
Directly replacing SUNCG textures with our non-tileable ones will create artifacts near boundaries. Most frameworks for tileable texture synthesis \cite{liang2001real,moritz2017texture} use randomized patch-based methods \cite{barnes2009patch}, which do not preserve structures such as sharp straight edges that are common for indoor scene materials such as bricks or wood floors. Instead, we first search for an optimal crop from our SVBRDF texture by minimizing gradients for diffuse albedo, normals and roughness perpendicular to the patch boundaries. We next find the best seam for tiling along the horizontal and vertical directions by modifying the graph cut method of \cite{kwatra2003graphcut} to encourage gradients to be similar at seams. Please refer to Appendix \ref{sec:textureSynthesis} for details on the energy design and examples of our texture synthesis.

\vspace{-0.4cm}
\paragraph{Step 2: Mapping SVBRDFs}

Once our materials are tileable, we must use them to replace SUNCG ones in a semantically meaningful way. Since the specular reflectance of SUNCG materials is not realistic, we do this only for diffuse textures and directly use specularity from our dataset to render images. We manually divide 633 most common diffuse textures from SUNCG and from our entire dataset into 10 categories based on appearance and semantic labels, such as fabric, stone or wood. We render both sets of diffuse textures on a planar surface under flash light and use an encoder network similar to \cite{li2018materials} to extract features, then use nearest neighbors to map the materials. We randomly choose from the $10$ pre-computed nearest neighbors to render images in our dataset.

\begin{figure}
\centering
\includegraphics[width = 3.2in]{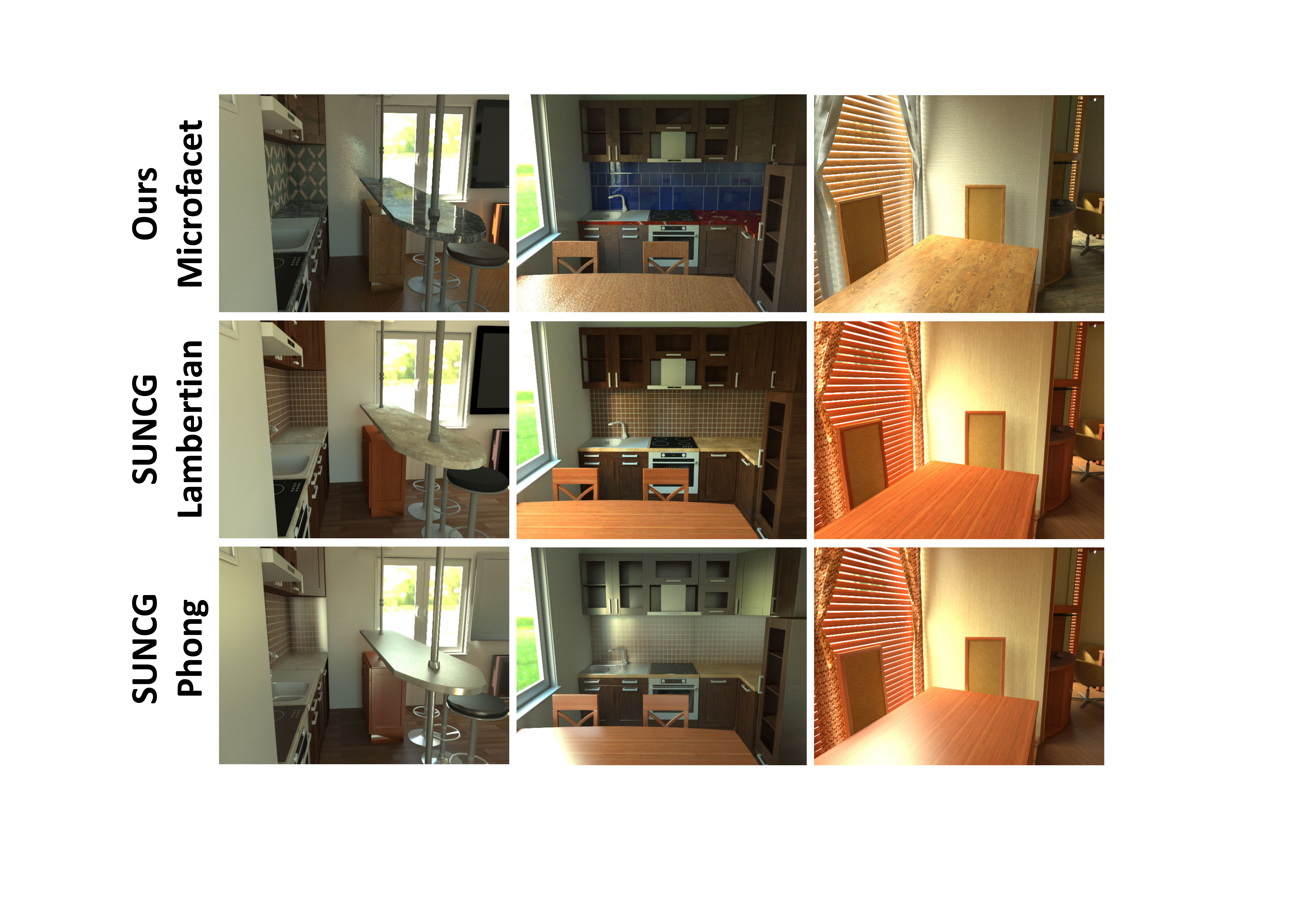}
\caption{The first row shows images rendered with materials from our dataset. The second and third rows are images rendered with the original materials from SUNCG dataset using Lambertian and Phong models. 
Images rendered with our materials have realistic specular highlights.}
\label{fig:mat_comparison}
\end{figure}

\vspace{-0.5cm}
\paragraph{Step 3: Mapping homogeneous BRDFs}
To map homogeneous materials from SUNCG to ours, we keep the diffuse albedo unchanged and map specular Phong parameters to our microfacet model. Since the two lobes are very different, a direct fitting does not work. Instead, we compute a distribution of microfacet parameters conditioned on Phong parameters based on the mapping of diffuse textures, then randomly sample from that distribution to map specular parameters. Specifically, let $\mathbf{x}_{P} \in \mathcal{P}$ be specular parameters of Phong model and $\mathbf{y}_{M} \in \mathcal{M}$ be those of our microfacet BRDF.
If a material from SUNCG has specular parameters $\mathbf{x}_{P} = \mathbf{p}_{b}$, we count the number of pixels in its 10 nearest neighbors from our dataset whose specular parameters are $\mathbf{y}_{M} = \mathbf{m}_{a}$. We sum up the number across the whole SUNCG dataset as $N(\mathbf{m}_{a}, \mathbf{p}_{b})$. The probability of material with specular parameters $\mathbf{y}_{M}$ given the original materials in SUNCG has specular parameters $\mathbf{x}_{P}$ is defined as: 
\begin{equation}
P(\mathbf{y}_{M} = \mathbf{m}_{a} | \mathbf{x}_{P} = \mathbf{p}_{b})  \nonumber
= \frac{N(\mathbf{p}_{b}, \mathbf{m}_{a} ) }{\sum_{\mathbf{m}_{c}\in \mathcal{M}}N(\mathbf{p}_{b}, \mathbf{m}_{c}) }.
\end{equation}
We sample the distribution as a piece-wise constant function and interpolate uniformly inside each bin to get continuous specular parameters of microfacet BRDF.

\vspace{-0.4cm}
\paragraph{Comparative results}
Figure \ref{fig:mat_comparison} shows a few scenes rendered with Lambertian, SUNCG Phong and our BRDF models.
Images rendered with Lambertian BRDF do not have any specularity, those with Phong BRDF have strong but flat specular highlights, while ours are clearly more realistic. All the materials in our rendering are perfectly tiled and assigned to the correct objects, which demonstrates the effectiveness of our material mapping pipeline.

\subsection{Spatially Varying Lighting}
To enable tasks such as object insertion or material editing, we must estimate lighting at every spatial location that encodes complex global interactions. Thus, our dataset must also render such ground truth. We do so by rendering a $16 \times 32$ environment map at the corresponding 3D point on object surfaces at every pixel. 

In Figure \ref{fig:lightApprox}, we show that an image obtained by integrating the product of this lighting and BRDF over the hemisphere looks very realistic, with high frequency specular highlights being correctly rendered. Note that global illumination and occlusion have already been baked into per pixel lighting, which makes it possible for a model trained on our lighting dataset to reason about those complex effects.

\vspace{-0.3cm}
\paragraph{Enhancing lighting variations}
The SUNCG dataset is rendered with only one outdoor environment map and two area light intensities ($400$ for light bulbs and $0.5$ for light shades). We add variations to the lighting to ensure generalizability, using $218$ HDR outdoor panoramas from \cite{holdgeoffroy19sky} and random RGB intensities for area lights.

\begin{figure}
\centering
\includegraphics[width=3.2in]{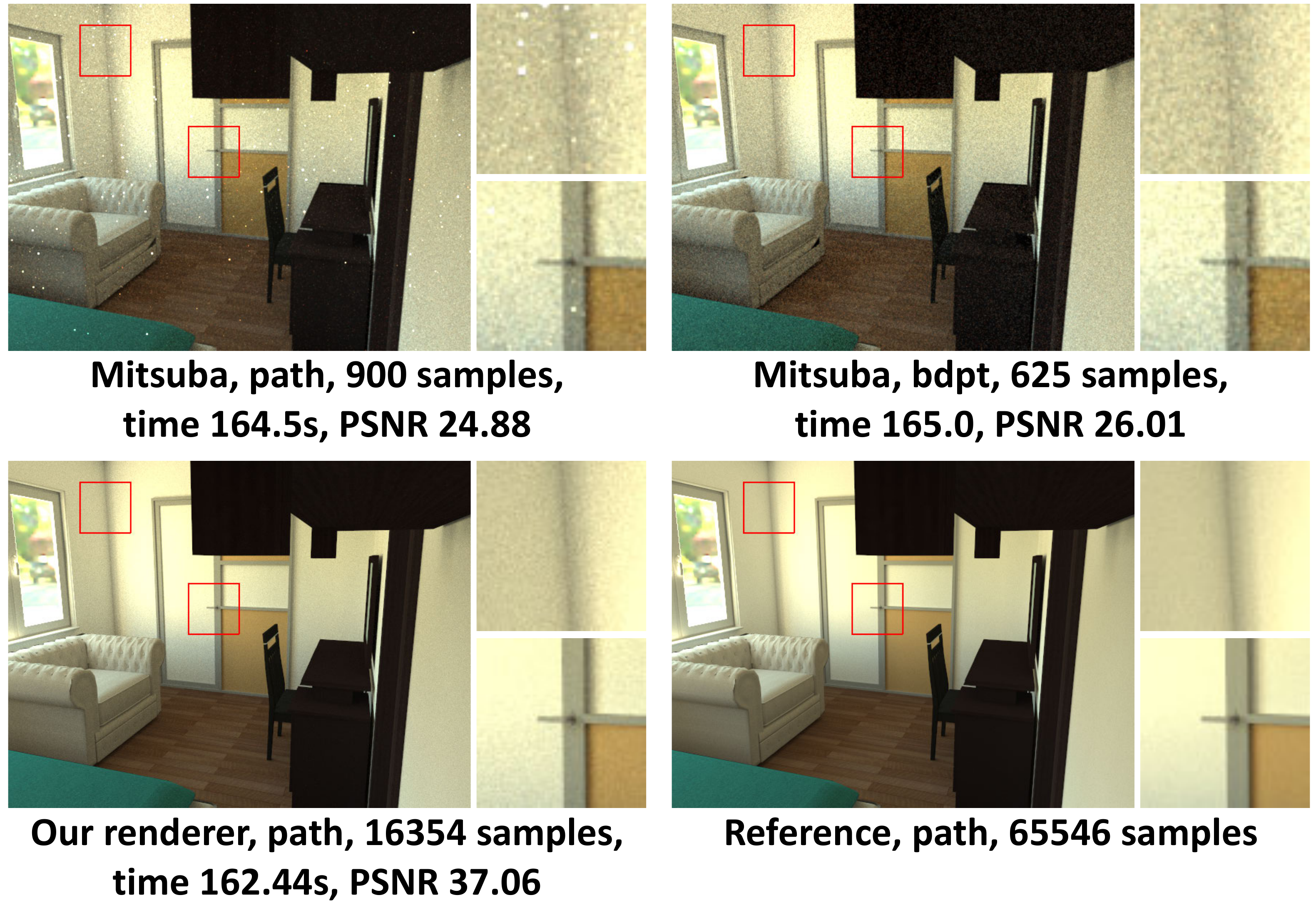}
\caption{Comparisons of images rendered with Mitsuba and our GPU renderer in the same amount of time using path tracing. The quality of the image rendered by our renderer in less than three minutes is much better. It takes about 50 minutes for Mitsuba to achieve similar results. }
\label{fig:renderer_comparison}
\end{figure}

\subsection{Fast Physically-Based Rendering}
To render high quality images with realistic appearances, it is necessary to use a physically based renderer that models complex light transport effects such as global illumination and soft shadows. However, current open source CPU renderers are too slow for creating a large dataset, especially to render per-pixel lighting. Thus, we implement our own physically-based GPU renderer using Nvidia OptiX \cite{OptiX}. To render a 480 $\times$ 640 image with 16384 samples per pixel, our renderer on Tesla V100 GPU needs 3-6 minutes, while Mitsuba on 16 cores of Intel i7-6900K CPU needs around $1$ hour. Figure \ref{fig:renderer_comparison} compares images rendered with Mitsuba \cite{Mitsuba} and with our renderer using the same amount of time. 

\vspace{-0.5cm}
\paragraph{Rendered dataset}
We render $78794$ HDR images, with $72220$ used for training and $6574$ for testing. The resolution of each image is $480\times 640$. We also render per pixel ground-truth lighting for $26719$ images in the training set and all images in the testing set, at a spatial resolution of $120\times 160$. Our dataset and renderer will be made publicly available.

\section{Network Design}
\label{sec::networkDesign}

\begin{figure*}
\includegraphics[width=6.7in]{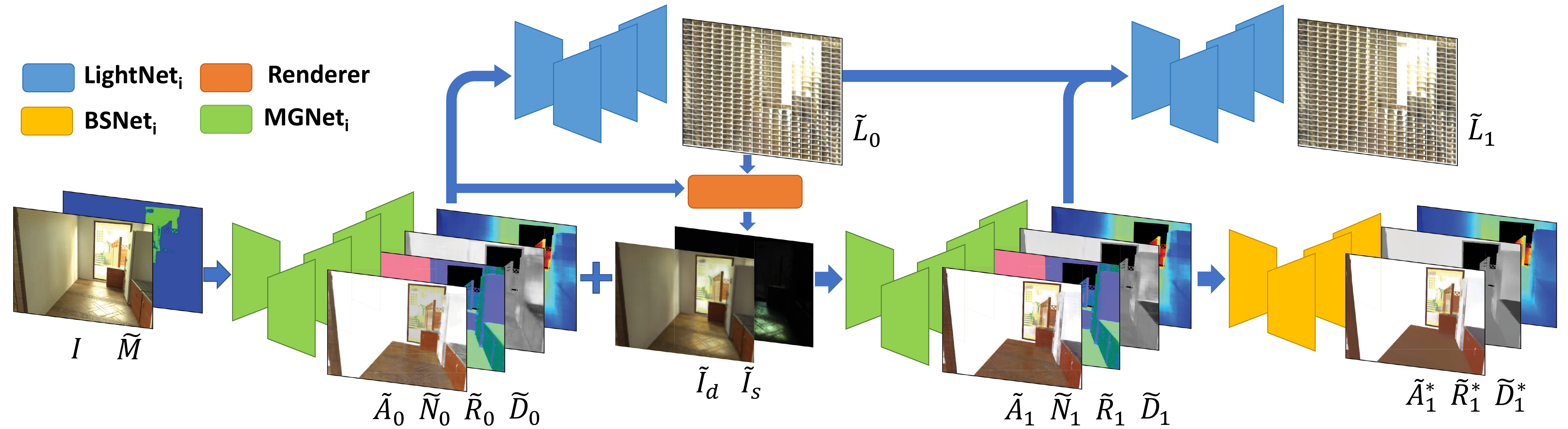}
\caption{Our network design consists of a cascade, with one encoder-decoder for material and geometry prediction and another one for spatially-varying lighting, along with a physically-based differentiable rendering layer and a bilateral solver for refinement.}
\label{fig:network}
\end{figure*}

\begin{figure}
\centering
\includegraphics[width=3.4in]{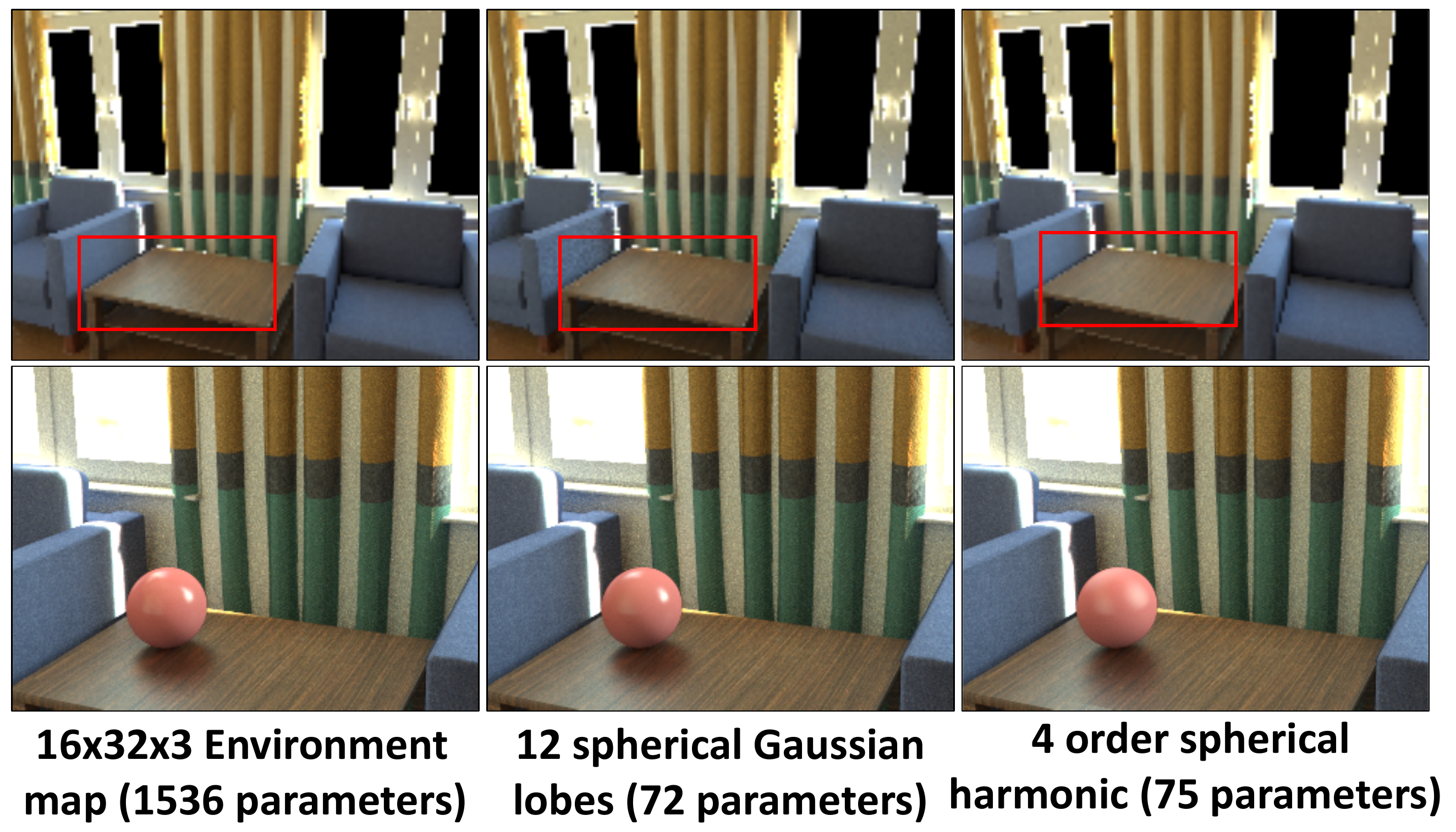}
\caption{Comparisons of images rendered with lighting approximations. The first row: images rendered by our rendering layer using ground-truth normals and materials but with different lighting representations. The second row: inserting a sphere into the scene. In both examples, we can clearly see that spherical Gaussians can recover high frequency lighting much better with fewer parameters.}
\label{fig:lightApprox}
\end{figure}

Estimating material, geometry and lighting from a single indoor image is an extremely ill-posed problem, which we solve using priors learned by our physically-motivated deep network (architecture shown in Figure \ref{fig:network}). Our network consists of cascaded stages of a SVBRDF and geometry predictor, a spatially-varying lighting predictor and a differentiable rendering layer, followed by a bilateral solver for refinement. 

\vspace{-0.4cm}
\paragraph{Material and geometry prediction}
The input to our network is a single gamma-corrected low dynamic range image $I$, stacked with a predicted three-channel segmentation mask $\{\tilde{M}_{o}, \tilde{M}_{a}, \tilde{M}_{e}\}$ that separates pixels of object, area lights and environment map, where ($\tilde{\cdot}$) represents predictions. The mask is obtained through a pre-trained network and useful since some predictions are not defined everywhere (for example, BRDF is not defined on light sources). Inspired by \cite{li2018materials,li2018learning}, we use a single encoder to capture correlations between material and shape parameters, obtained using four decoders for diffuse albedo ($A$), roughness ($R$), normal ($N$) and depth ($D$). Skip links are used for preserving details. Then the initial estimates of material and geometry are given by
\begin{equation}
    \Tilde{A}, \Tilde{N}, \Tilde{R}, \Tilde{D} = \mathbf{MGNet}_{0}(I, M).
\end{equation}

\vspace{-0.4cm}
\paragraph{Spatially Varying Lighting Prediction}
Inverse rendering for indoor scenes requires predicting spatially varying lighting for every pixel in the image. Using an environment map as the lighting representation leads to a very high dimensional output space, that causes memory issues and unstable training due to small batch sizes. Spherical harmonics are a compact lighting representation that have been used in recent works \cite{kanamori2018relighting,li2018learning}, but do not efficiently recover high frequency lighting necessary to handle specular effects \cite{ramamoorthi2001efficient,basri2003lambertian}. Instead, we follow pre-computed radiance transfer methods \cite{tsai2006all,green2007efficient,xu2013anisotropic} and use isotropic spherical Gaussians that approximate all-frequency lighting with a smaller number of parameters. We model the lighting as a spherical function $L(\eta)$ approximated by the sum of spherical Gaussian lobes:
\begin{equation}
L(\eta) = \sum_{k=1}^{K}F_{k}G(\eta; \xi_{k}, \lambda_{k} ), \;\; G(\eta; \xi, \lambda) = e^{\lambda(1 - \eta\cdot \xi)},
\label{eq:SGapproxEnv}
\end{equation}
where $\eta$ and $\xi$ are vectors on the unit sphere $\mathcal{S}^{2}$, $F_{k}$ controls RGB color intensity and $\lambda$ controls the bandwidth.

Each spherical Gaussian lobe is represented by 6 parameters  $\{\xi_{k}, \lambda_{k}, F_{k}\}$.
Figure \ref{fig:lightApprox} compares the images rendered with a 12-spherical Gaussian lobes approximation (72 parameters) and a fourth-order spherical harmonics approximation (75 parameters). Quantitative comparisons of lighting approximation and rendering errors are in Table \ref{tab:lightApprox}. It is evident that even using fewer parameters, the spherical Gaussian lighting performs better, especially close to specular regions.

Our novel lighting prediction network, $\mathbf{LightNet}_{0}(\cdot)$, accepts predicted material and geometry as input, along with the image. It uses a shared encoder and separate decoders with a $\texttt{tanh}$ layer to predict:
\begin{equation}
\{\bar{\xi}_{k}\}, \{\bar{\lambda}_{k} \}, \{\bar{F}_{k} \} =  \mathbf{LightNet}_{0}(I, \tilde{M}, \tilde{A}, \tilde{N}, \tilde{R}, \tilde{D}).
\label{eq:lightNet}
\end{equation}

These original low dynamic range parameters are mapped to high dynamic range parameters  $\{\tilde{\xi}_{k}\}$, $\{\tilde{\lambda}_{k}\}$ and $\{\tilde{F}_{k}\}$ through following non-linear transformation. 
\begin{eqnarray}
\tilde{\xi}_{k} &=& \frac{\bar{\xi}_{k} }{||\bar{\xi}_{k}||_{2}^{2} }\\
\tilde{\lambda}_{k} &=& \tan\left(\frac{\pi}{4}(\bar{\lambda}_{k}+1)\right) \\
\tilde{F}_{k} &=& \tan\left(\frac{\pi}{4}(\bar{F}_{k}+1)\right).
\end{eqnarray}

Thus, our final predicted lighting is HDR, which is important for applications like relighting and material editing.

\begin{table}
\centering
\footnotesize
\begin{tabular}{|l|c|c|}
\hline
& Lighting ($\log L_2$)  & Image ($L_2$)   \\
\hline
SH (75 para.)  &4.43  & $8.6\times 10^{-3}$\\
\hline 
SG (72 para.) & 1.56  & $7.6\times 10^{-3}$\\
\hline
\end{tabular}
\vspace{0.1cm}
\caption{Quantitative comparison of using spherical harmonic (SH) and spherical Gaussian (SG) for lighting representation. From left to the right, the average error when using each representation to approximate per pixel lighting in Figure \ref{fig:lightApprox}, the MSE of the rendered images. Again, spherical Gaussian performs better.}
\label{tab:lightApprox}
\end{table}

\vspace{-0.4cm}
\paragraph{Differentiable rendering layer}
Our dataset in Section \ref{sec::dataset} provides ground truth for all scene components. But to model realistic indoor scene appearance, we additionally use a differentiable in-network rendering layer to mimic the image formation process, thereby weighting those components in a physically meaningful way. We implement this layer by numerically integrating the product of SVBRDF and spatially-varying lighting over the hemisphere. 
Let $l_{ij} = l(\phi_{i}, \theta_{j})$ be a set of light directions sampled over the upper hemisphere, and $v$ be the view direction. The rendering layer computes the diffuse image $\tilde{I}_{d}$ and specular image $\tilde{I}_s$ as: 
{\small
  \begin{align}
\tilde{I}_{d} &= \sum_{i,j} f_{d}(v, l_{ij}; \tilde{A}, \tilde{N}) L\left(l_{ij}; \{\xi_{k}, \lambda_{k}, F_k\}\right)\cos \theta_{j} \text{d}\omega ,
\label{eq:diffIm}\\
\tilde{I}_{s} &= \sum_{i,j} f_{s}(v, l_{ij}; \tilde{R}, \tilde{N}) L\left(l_{ij}; \{\xi_{k}, \lambda_{k}, F_k\}\right)\cos \theta_{j} \text{d}\omega ,
\label{eq:specIm}
  \end{align}
}
where $\text{d}\omega$ is the differential solid angle.
We sample $16 \times 8$ lighting directions. While this is relatively low resolution, we empirically find, as shown in Figure \ref{fig:lightApprox}, that it is sufficient to recover most high frequency lighting effects. 

\vspace{-0.4cm}
\paragraph{Loss Functions}
Our loss functions incorporate physical insights.
We first observe that two ambiguities are difficult to resolve: the ambiguity between color and light intensity, as well as the scale ambiguity of single image depth estimation. Thus, we allow the related loss functions to be scale invariant. For material and geometry, we use the scale invariant $L_2$ loss for diffuse albedo ($\mathcal{L}_{A}$), $L_2$ loss for normal ($\mathcal{L}_{N}$) and roughness ($\mathcal{L}_{R}$) and a scale invariant $\log$-encoded loss for depth ($\mathcal{L}(D)$) due to its high dynamic range:
\begin{equation}
\mathcal{L}_{D} = \lVert (\log(D+1) - \log(c_{d}\tilde{D} + 1))\odot (M_{a} + M_{o}) \rVert_{2}^{2} ,
\label{eq:lossDepth}
\end{equation}
where $c_{d}$ is a scale factor computed by least squares regression. 
For lighting estimation, we find supervising both the environment maps and spherical Gaussian parameters is important for preserving high frequency details. Thus, we compute ground-truth spherical Gaussian lobe parameters by approximating the ground-truth lighting using the LBFGS method\footnote{Please refer to Appendix \ref{sec:ground-truthSphericalGaussianLobes} for details on how we compute ground-truth spherical Gaussian parameters.}. We use the same scale invariant $\log$-encoded loss as \eqref{eq:lossDepth} for weights ($\{\mathcal{L}_{F_{k}}\}$), bandwidth ($\{\mathcal{L}_{\lambda_{k}}\}$) and lighting ($\{\mathcal{L}_{L}\}$), with an $L_2$ loss for direction $(L_{\xi_{k}})$. We also add a a scale invariant $L_2$ rendering loss:
\begin{equation}
\mathcal{L}_{ren} = ||(I - c_{diff}\tilde{I}_{d} - c_{spec}I_{s}) \odot M_{o}||_{2}^{2}
\end{equation}
where $\tilde{I}_{d}$ and $\tilde{I}_{s}$ are rendered using \eqref{eq:diffIm} and \eqref{eq:specIm}, respectively, while $c_{diff}$ and $c_{spec}$ are positive scale factors computed using least square regression. The final loss function is a weighted summation of the proposed losses:
\begin{eqnarray}
\mathcal{L}\!\!\!\!&=&\!\!\!\!\alpha_{A}\mathcal{L}_{A} + \alpha_{N}\mathcal{L}_{N} + \alpha_{R}\mathcal{L}_{R} + \alpha_{D}\mathcal{L}_{D} + \alpha_{L}\mathcal{L}_{L} \nonumber \\
\!\!\!\!&&\!\!\!\!\alpha_{ren}\mathcal{L}_{ren} + \sum_{k=1}^{K}\alpha_{\lambda}\mathcal{L}_{\lambda_{k}} + \alpha_{\xi}\mathcal{L}_{\xi_{k}} + \alpha_{F}\mathcal{L}_{F_{k}} .
\label{eq:loss}
\end{eqnarray}

\vspace{-0.4cm}
\paragraph{Refinement using bilateral solver}
We use an end-to-end trainable bilateral solver to impose a smoothness prior \cite{barron2016fast,li2018cgintrinsics}. The inputs to a bilateral solver include the prediction, the estimated diffuse albedo $\tilde{A}$ as a guidance image, and confidence map $C$.
We train a shallow network with three sixteen-channel layers for confidence map predictions. 
Let $\mathbf{BS}_{X}(\cdot)$ be the bilateral solver and $\mathbf{BSNet}_{X}(\cdot)$ be the network for confidence map predictions where $X \in \{A, R, D\}$. We do not find refinement to have much effect on normals.
The refinement process is: 
\begin{eqnarray}
\tilde{C}_{X}\!\!\!\!&=&\!\!\!\!\mathbf{BSNet}_{X}(\tilde{X}, I, \tilde{M}), \quad X \in \{A, R, D\} \\
\tilde{X}^{*}\!\!\!\!&=&\!\!\!\! \mathbf{BS}_{X}(\tilde{X}; C_{X}, \tilde{A})
\end{eqnarray}
where we use $(^{*})$ for predictions after refinement. 

\vspace{-0.4cm}
\paragraph{Cascade Network}
Akin to recent works on high resolution image synthesis \cite{karras2017progressive,chen2017photographic} and inverse rendering \cite{li2018learning}, we introduce a cascaded network that progressively increases resolution and iteratively refines the predictions through global reasoning. We achieve this by sending both the predictions and the rendering layer applied on the predictions to the next cascade stages, $\mathbf{MGNet}_{1}(\cdot)$ for material and geometry and $\mathbf{LightNet}_{1}(\cdot)$ for lighting, so that the network can reason about their differences. 
\begin{eqnarray}
\tilde{A}_{1}, \tilde{N}_{1} \tilde{R}_{1}, \tilde{D}_{1} \!\!\!\!&=&\!\!\!\! \mathbf{MGNet}_{1}(I, \tilde{M}, \tilde{A}_0, \tilde{N}_0, \tilde{R}_{0}, \tilde{D}_{0}, \nonumber \\
&&  \qquad c_{diff}\tilde{I}_d, c_{spec}\tilde{I}_s )  \\
\{\bar{\xi}_{k}\}_{1}, \{\bar{\lambda}_{k}\}_{1}, \{\bar{F}_{k}\}_{1} \!\!\!\!&=&\!\!\!\! \mathbf{LightNet}_{1}(I, \tilde{M}, \tilde{A}_1, \tilde{N}_1, \tilde{R}_1, \nonumber \\
&& \tilde{D}_1, \{\bar{\xi}_{k}\}_{0}, \{\bar{\lambda}_{k}\}_{0}, \{\bar{F}_{k}\}_{0})
\end{eqnarray}

Cascade stages have similar architectures as their initial network counterparts. One thing to notice is that we send low dynamic range lighting predictions $\{\bar{\xi}_{k}\}_{0}$, $\{\bar{\lambda}_{k}\}_{0}$, $\{\bar{F}_{k}\}_{0}$ instead of the high dynamic range predictions, because we observe that it makes training more stable.

\vspace{-0.4cm}
\paragraph{Training Details}
It is hard to train our whole pipeline end-to-end from scratch due to limited GPU memory, even with the use of group normalization \cite{wu2018group}.
So, we first train $\mathbf{MGNet}_{i}(\cdot)$ and $\mathbf{LightNet}_i(\cdot)$ separately with large batch sizes, fine-tune them together with smaller batch sizes, and finally train the bilateral solver.  Please refer to Appendix \ref{sec:trainingDetails} for training details and hyperparameter choices.

\section{Experiments}
\label{sec::experiments}
\begin{figure*}[h]
\centering
\includegraphics[width=1.0\linewidth]{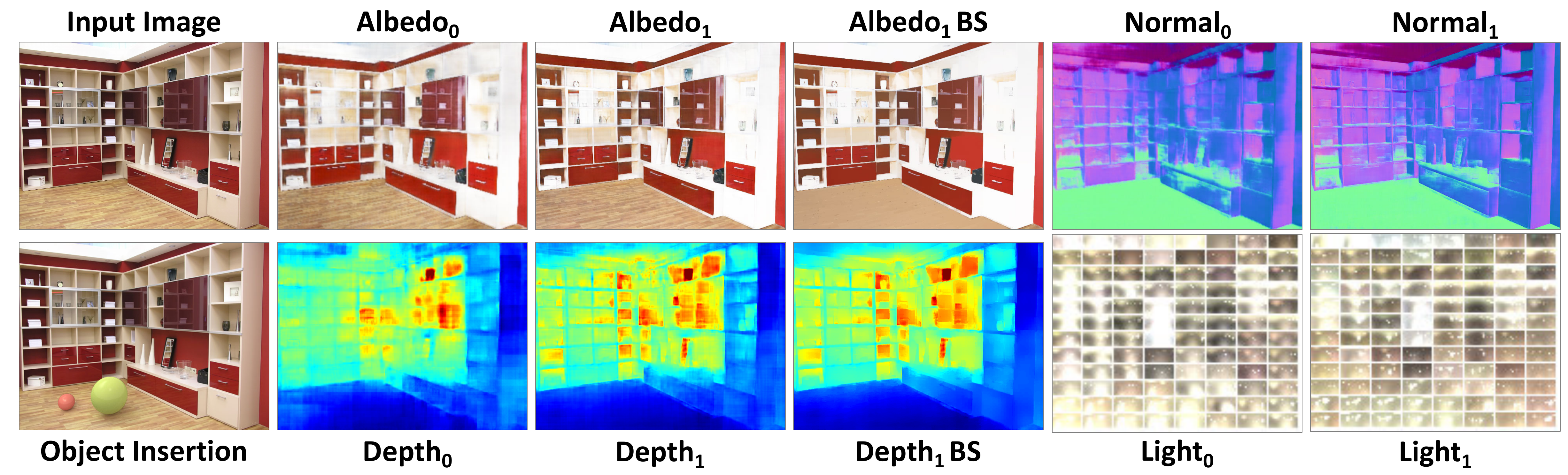}
\caption{Impact of cascade and bilateral solver on a real example. Improvements are observed due to the cascade structure and bilateral solver. The estimates are accurate enough to insert a novel object with realistic global illumination effects.}
\label{fig:cascadeReal}
\end{figure*}

\begin{figure*}[!!t]
\centering
\includegraphics[width=1.0\linewidth]{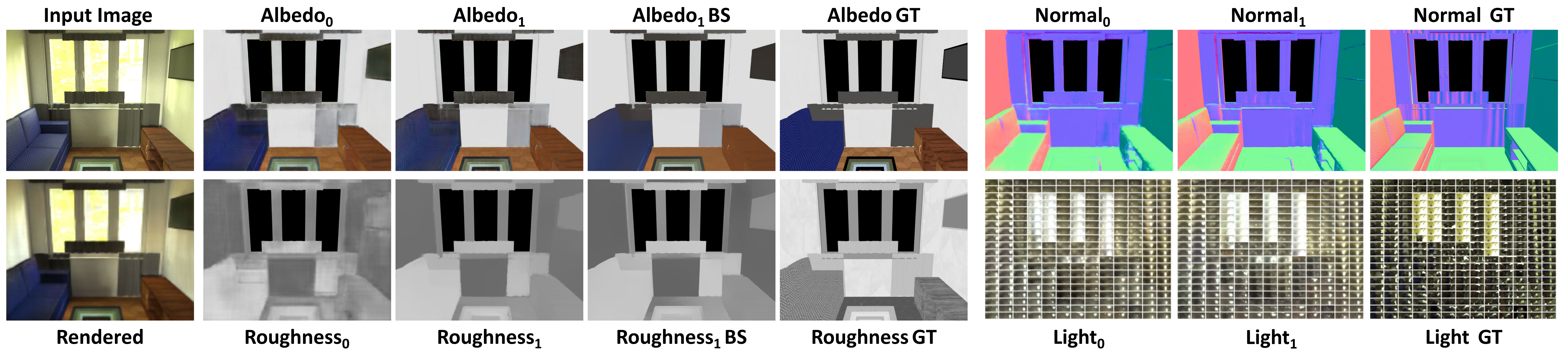}
\caption{Impact of cascade and bilateral solver on a synthetic example. We observe that all the scene components benefit from cascaded estimation, while the bilateral solver is effective at refinement. The output of the rendering layer closely matches the input.}
\label{fig:cascadeSynthetic}
\end{figure*}

\begin{figure*}[!!t]
\centering
\includegraphics[width=1.0\linewidth]{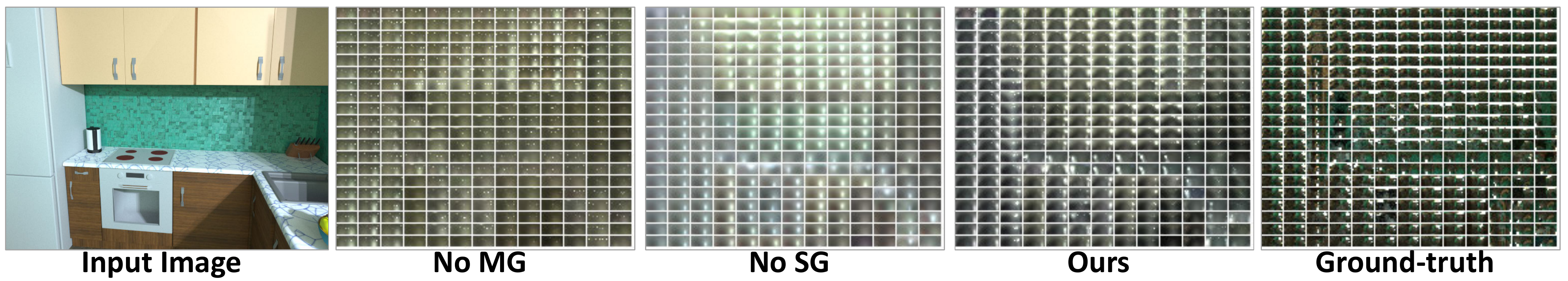}
\caption{Comparison of lighting predictions. From left to the right are input image, \textbf{No MG: }without predicted material and geometry as input, \textbf{No SG: }without ground-truth spherical Gaussian parameters as supervision and our predictions and the ground-truth lighting. }
\label{fig:ablation_lighting}
\end{figure*}

\begin{figure*}
\centering
\includegraphics[width=1.0\linewidth]{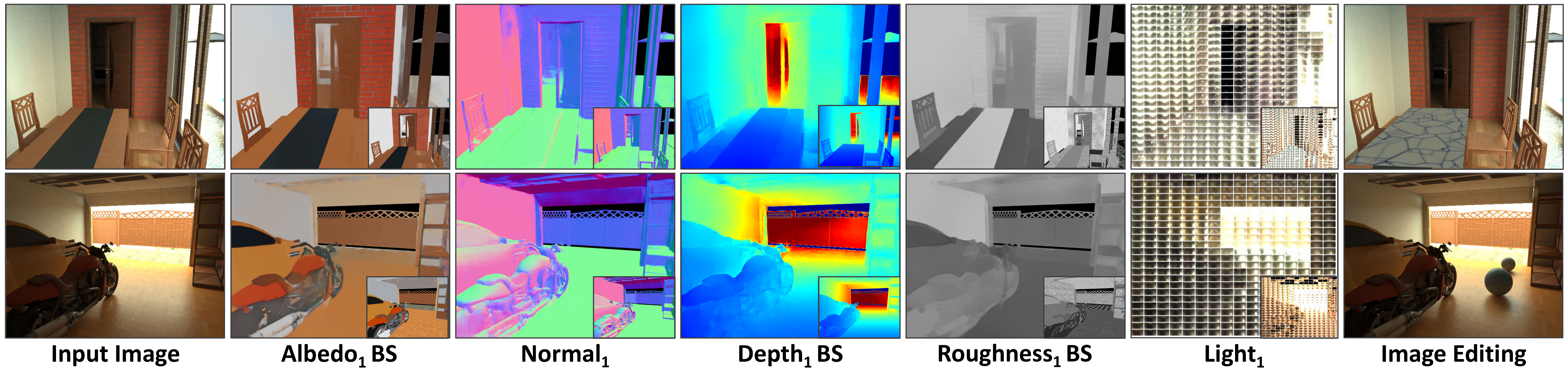}
\caption{Results on our synthetic dataset. Given an input image, our estimated albedo, normals, depth, roughness and lighting are close to ground truth shown as insets. These are used for material editing (top) and object insertion (bottom).}
\label{fig:synthetic}
\end{figure*}

Our experiments highlight the effectiveness of our dataset and network for single image inverse rendering in indoor scenes, through shape, material and lighting estimation. We achieve high accuracy on synthetic data and competitive performance on real images with respect to methods that focus only on a subset of those tasks. We conduct studies on the roles of various components in our pipeline. Finally, we illustrate applications such as high quality object insertion and material editing in real images that can only be enabled by our holistic solution to inverse rendering.
               
\subsection{Analysis of Network and Training Choices}
We study the effect of the cascade structure, joint training and refinement. Quantitative results for material and geometry predictions on the proposed dataset are summarized in Table \ref{tab:MGNet}, while those for lighting are shown in Table \ref{tab:LightNet}.

\vspace{-0.4cm}
\paragraph{Cascade}
The cascade structure leads to clear gains for shape, BRDF and lighting estimation by iteratively improving and upsampling our predictions in Tables \ref{tab:MGNet} and \ref{tab:LightNet}. This holds for both real data and synthetic data, as shown in Figure \ref{fig:cascadeReal} and Figure \ref{fig:cascadeSynthetic}. We observe that the cascade structure can effectively remove noise and preserve high frequency details for both materials and lighting. The errors in our shape, material and lighting estimates are low enough to photorealistically edit the scene to insert new objects, while preserving global illumination effects. In Figure \ref{fig:cascadeSynthetic}, we observe that the image rendered using our predicted material, shape and lighting closely match the input image. 

\vspace{-0.4cm}
\paragraph{Joint training for inverse rendering}
Next we study whether BRDF, shape and lighting predictions can help improve each other. We compare jointly training the whole pipeline (``Joint'') using the loss in \eqref{eq:loss} and compare to independently training (``Ind'') each component $\mathbf{MGNet}_{i}$ and $\mathbf{LightNet}_{i}$. Quantitative errors on Tables \ref{tab:MGNet} and \ref{tab:LightNet} show that while errors for shape and BRDF prediction remain similar, those for rendering and lighting decrease. Next, we test lighting predictions without predicted BRDF as input for the first level of cascade (``No MG''). Both quantitative results in Table \ref{tab:LightNet} and qualitative comparison in Figure \ref{fig:ablation_lighting} demonstrate that the predicted BRDF and shape are important for the network to recover spatially varying lighting.  We can see without the predicted material and geometry as input, the predicted lighting---especially the ambient color---does not sufficiently adapt spatially to the scene (possibly because of ambiguities between lighting and surface reflectance). This justifies our choice of jointly reasoning about  shape, material and lighting. We also test lighting predictions with and without ground-truth SVSG parameters as supervision (``No SG''), finding that direct supervision leads to a sharper lighting prediction, which is shown in Figure \ref{fig:ablation_lighting}.

\vspace{-0.4cm}
\paragraph{Refinement}
Finally, we study the impact of the bilateral solver. Quantitative improvements over the second cascade stage in Table \ref{tab:MGNet} are modest, which indicates that the network already learns good smoothness priors by that stage. This is shown in Figure \ref{fig:cascadeSynthetic}, where the second level of cascade network generates smooth predictions for both material and lighting. But we find the qualitative impact of the bilateral solver to be noticeable on real images (for example, diffuse albedo in Figure \ref{fig:cascadeReal}), thus, we use it in all our real experiments.

\vspace{-0.4cm}
\paragraph{Qualitative examples}
In Figure \ref{fig:synthetic}, we use a single input image from our synthetic test set to demonstrate depth, normal, SVBRDF and spatially-varying lighting estimation. The effectiveness is illustrated by low errors with respect to ground truth. Accurate shading and global illumination effects on an inserted object, as well as photorealistic editing of scene materials, show the utility of our decomposition.

\begin{table}[t]
\footnotesize
\centering
\renewcommand{\arraystretch}{1.2}
\begin{tabular}{|l|c|c|c|c|c|}
\hline
& \multicolumn{2}{|c|}{Cascade 0} & \multicolumn{3}{|c|}{Cascade 1} \\
\hline 
& Ind. & Joint & Ind. & Joint & BS \\ 
\hline
$A (10^{-2})$ & 1.28& 1.28   &1.18  &1.18  &\textbf{1.16}  \\
\hline 
$N (10^{-2})$ & 4.91& 4.91  &4.91   &4.51  &\textbf{4.51}  \\
\hline 
$R (10^{-1})$ & 1.72 & 1.72   &1.72  &1.72  & \textbf{1.70}  \\
\hline
$D (10^{-2})$ & 8.06 & 8.00 & 7.29  &7.26  &\textbf{7.20}  \\
\hline
\end{tabular}
\vspace{0.1cm}
\caption{Quantitative comparisons of shape and material reconstructions on our test set. We use scale invariant L2 error for diffuse albedo ($A$), scale invariant $\log^{2}$ error for depth ($D$) and L2 error for normal ($N$) and roughness ($R$).}
\label{tab:MGNet}
\end{table}

\begin{table}[t]
\footnotesize
\centering
\renewcommand{\arraystretch}{1.2}
\begin{tabular}{|l|c|c|c|c|c|c|}
\hline
& \multicolumn{4}{|c|}{Cascade 0} & \multicolumn{2}{|c|}{Cascade 1} \\
\hline 
& No MG & No SG & Ind. & Joint & Ind. & Joint \\ 
\hline
$L $ & 2.83 &2.85  &2.54  &2.50  &2.49  &2.43  \\
\hline 
$I (10^{-2})$ & 5.00 &1.56  &1.56  &1.06  &1.92  &1.11  \\
\hline
\end{tabular}
\vspace{0.1cm}
\caption{Quantitative comparison of lighting predictions on  test set. We use scale invariant L2 error for rendered image ($I$) and scale invariant $\log^{2}$ error for lighting ($L$).}
\label{tab:LightNet}
\end{table}

\subsection{Comparisons with Previous Works}
We address the problem of holistic inverse rendering with spatially-varying material and lighting which has not been tackled earlier. Yet, it is instructive to compare our approach to prior ones that focus on specific sub-problems.

\begin{table}[t]
\footnotesize
\centering
\begin{tabular}{|c|c|c|}
\hline 
Method & Training Set & WHDR \\ 
\hline
Ours (cascade 0)  & Ours  & 23.29 \\
\hline
Ours (cascade 1)  & Ours  & 21.99 \\
\hline
Ours (cascade 0)  & Ours + IIW  & 16.83  \\
\hline
Ours (cascade 1)  & Ours + IIW & 15.93 \\
\hline
Li. et al\cite{li2018cgintrinsics} & CGI + IIW & 17.5 \\
\hline
\end{tabular}
\vspace{0.1cm}
\caption{Intrinsic decomposition on the IIW dataset. Lower is better for the WHDR metric used here. }
\label{tab:IIW}
\end{table}

\vspace{-0.4cm}
\paragraph{Intrinsic decomposition} We compare two versions of our method on the IIW dataset \cite{bell2014intrinsic} for intrinsic decomposition evaluation: our network trained on our data alone and our network fine-tuned on the IIW dataset. The results are tabulated in Table \ref{tab:IIW}. We observe that the cascade structure is beneficial. We also observe a lower error compared to the prior work of \cite{li2018cgintrinsics}, which indicates the benefit of our dataset that is rendered with a higher photorealism, as well as a network design that closely reflects physical image formation.

\begin{figure}
\centering
\includegraphics[width=3.2in]{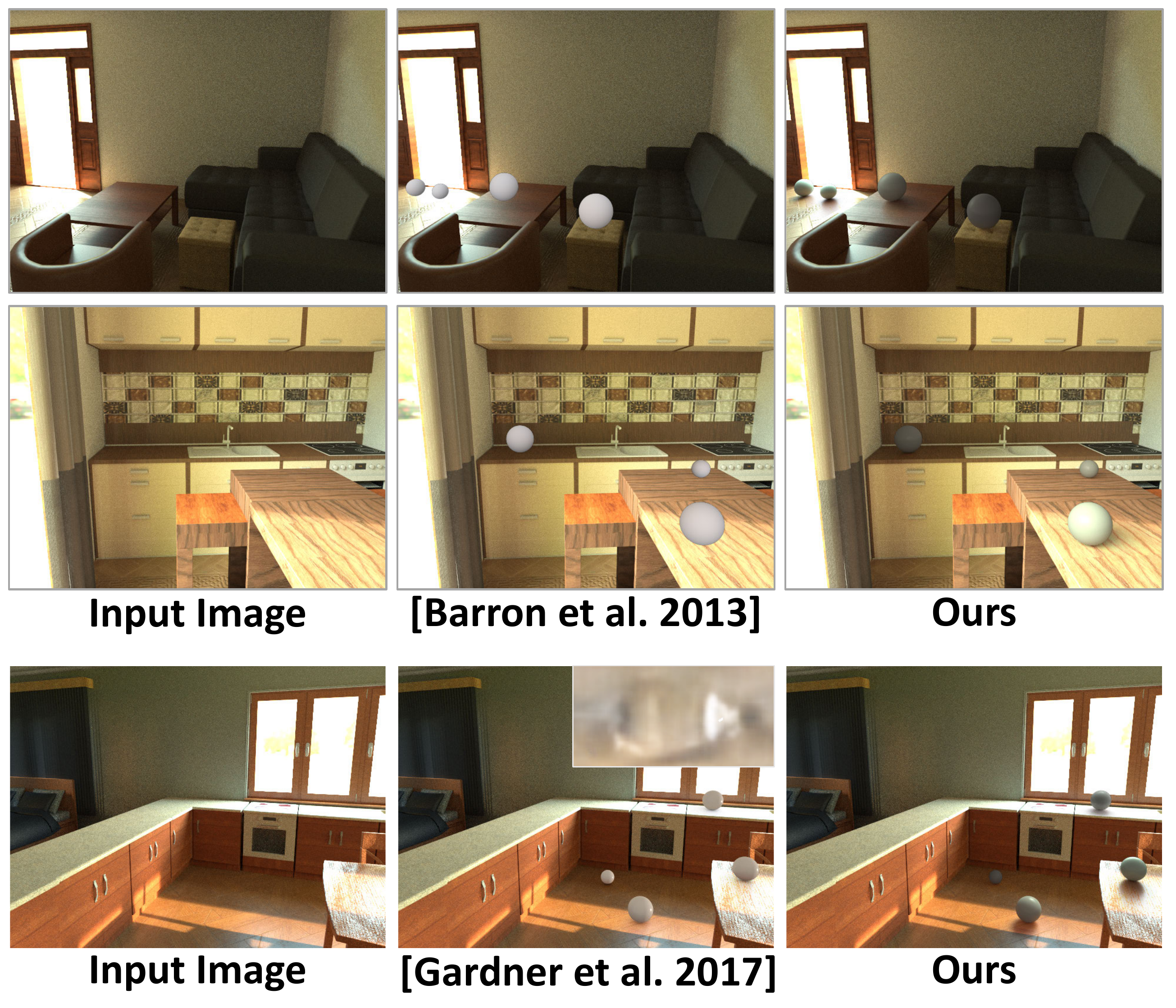}
\caption{Comparison of object insertion with Barron et al.~\cite{barron2013intrinsic} (First two rows) and Gardner et al. \cite{gardner2017indoor} (third rows) on synthetic examples. We observe that rendered appearances from \cite{barron2013intrinsic} are flat, while ours reflect the spatial variation in lighting at different parts of the scene. While method of \cite{gardner2017indoor} can preserve high frequency in the lighting, their lighting directions are not accurate and they could not handle spatially varying lighting. }
\label{fig:compare_lighting}
\end{figure}

\vspace{-0.4cm}
\paragraph{Lighting estimation}
We first compare to the method of Barron et al.~\cite{barron2013intrinsic} on our test set. Our scale-invariant shading errors on \{R, G, B\} channels are $\{0.87, 0.86, 0.83\}$, compared to their $\{2.33, 2.10, 1.90\}$. Our shape, material and spatially-varying lighting estimation, together with a physically-motivated network trained on a realistic large-scale dataset, lead to this large improvement. Qualitative comparisons are shown in Figure \ref{fig:compare_lighting}, where we render specular spheres into the image using different lighting predictions. While our method can clearly capture the complex lighting variations and high frequency components, the spheres rendered with the predicted lighting of \cite{barron2013intrinsic} are diffuse and have similar intensity across different regions of the image. Since only spherical harmonics parameters for log shading are predicted by \cite{barron2013intrinsic}, there is no physically correct way to turn its estimated spherical harmonics into environment lighting. Therefore, it cannot handle shadows and inter-reflections between object and the scene. Further, since only two orders of spherical harmonic parameters are predicted by \cite{barron2013intrinsic}, it cannot handle high frequency lighting. 

Next, we also compare with the work of Gardner et al.~\cite{gardner2017indoor}, which predicts a single environment lighting for the whole indoor scene. Quantitative results on our test set show that their mean $\log$ $L_2$ error across the whole image is 3.34 while our $\log$ $L_2$ error is 2.43. Qualitative results are shown in Figure \ref{fig:compare_lighting}. Since only one lighting for the whole scene is predicted by \cite{gardner2017indoor}, no spatially-varying lighting effects can be observed. 

\begin{figure*}
\centering
\includegraphics[width = 0.98\linewidth]{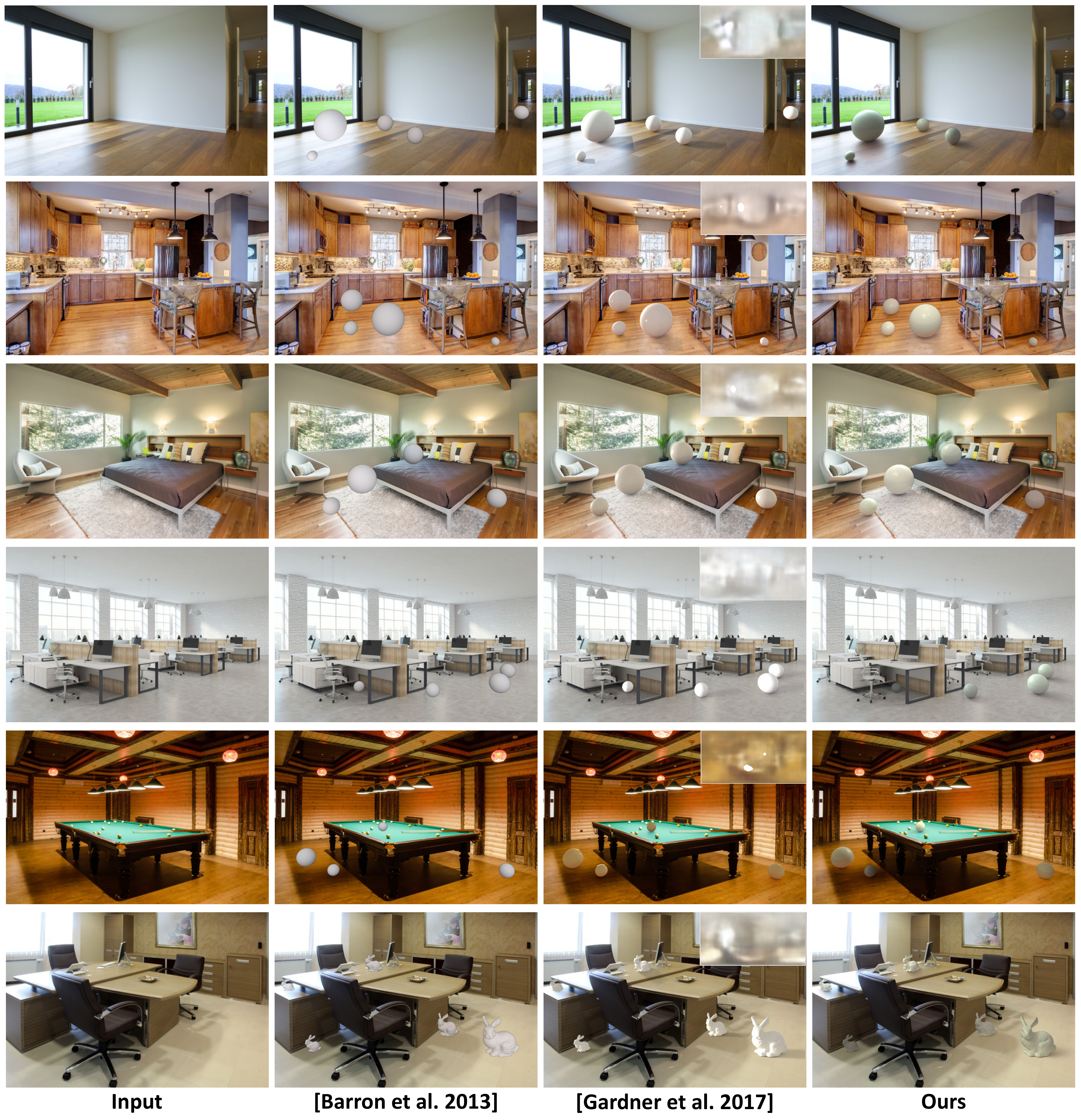}
\caption{\small Object insertion examples and comparisons. Our proposed method estimates shape (depth and surface normals), spatially-varying complex reflectance (based on a micro-facet SVBRDF model) and spatially-varying lighting from a single image of an indoor scene. Given these estimates, we can insert virtual 3D objects into these images and produce photo-realistic results where the objects look like they truly belong in the scene. Note the shading and specular highlights on the inserted spheres, the realistic shadows cast on the ground, the reflections from the ground onto the spheres and adaptation of the  appearance of the spheres to the local shadows and shading in the scene. 
We also compare with previous works of Barron et al.~\cite{barron2013intrinsic} and Gardner et al.~\cite{gardner2017indoor} on real images. Note that in the results of \cite{gardner2017indoor}, the shadows of some objects might be truncated by the plane we segment from the scene. 
}
\label{fig:object_insertion}
\end{figure*}

In Figure \ref{fig:object_insertion}, we compare our method with \cite{barron13sirfs} and \cite{gardner2017indoor} on several real examples for object insertion in an image with spatially-varying illumination. It is clear that our method achieves a significant improvement in object insertion. 

\begin{table}[]
\small
\centering
\begin{tabular}{|c|c|c|c|c|}
\hline
Method  & Mean($^\circ$) & Median($^\circ$) & Depth(Inv.)  \\
\hline
Ours (cascade 0) &27.08  & 21.14  & 0.217\\
\hline
Ours (cascade 1) &26.33  &20.21  & 0.206 \\
\hline
\end{tabular}
\vspace{0.1cm}
\caption{Normal and depth estimation on NYU dataset. }
\label{tab:NYU}
\end{table}

\vspace{-0.4cm}
\paragraph{Depth and normal estimation} We fine-tune our network, trained on our synthetic dataset, on NYU dataset as discussed in Appendix \ref{sec:trainingDetails}. The test error on NYU dataset is summarized in Table \ref{tab:NYU}. When testing depth error, we do not consider ground-truth depth value smaller than 1 or larger than 10 since they are outside the valid range of the sensor. When testing normal error, we mask out regions without accurate ground-truth normals. For both depth and normal prediction, the cascade structure consistently helps improve performance. Zhang et al.~\cite{zhang2016physically} achieve state-of-the-art performance for normal estimation using a more complex fine-tuning strategy by choosing images with similar appearance as NYU dataset and with more than six times as much training data. Eigen et al.~\cite{eigen2015depth} achieve better results by using 120K frames of raw video data to train their network, while we pre-train on synthetic images with larger domain gap, using only use 795 images from NYU dataset for fine-tuning. Although we do not achieve competitive performance on this task, it's not our main focus. Rather, we illustrate the wide utility of our proposed dataset and demonstrate estimation of factors of image formation good enough to support photorealistic augmented reality applications.

\subsection{Novel Applications}
Learning a disentangled shape, SVBRDF and spatially-varying lighting representation allows new applications that were hitherto not possible. We consider two of them here, object insertion and material editing. Before we discuss the two applications, we first describe how we resolve the ambiguity between scales of lighting and diffuse albedo.

\vspace{-0.4cm}
\paragraph{Scales of lighting and diffuse albedo}
We use scale invariant loss for both diffuse albedo and lighting prediction. However, for real applications, we need to recover the scale of both diffuse albedo and lighting. Let $c_{a}$ and $c_{l}$ be the coefficients of diffuse albedo and lighting, respectively. Recall that our rendering layer outputs a diffuse image $\tilde{I}_{d}$ and a specular image $\tilde{I}_{s}$. We can compute coefficients $c_{d}$  and $c_{s}$ to minimize the $L_2$ error between $c_{d}\tilde{I}_{d} + c_{s}\tilde{I}_{s}$ and input image $I$. Since our specular albedo is a constant, the scaling factor for our lighting prediction will be $c_{s}$ and we have
\begin{eqnarray}
c_{l} &=& c_{s} \label{eq:scale_s11}\\
c_{a} &=& \frac{c_d}{c_l} \label{eq:scale_s12}
\end{eqnarray}
However, for some images, specularity might be hard to observe, in which case we neglect the specularity term and simply compute the coefficient using 
\begin{eqnarray}
c_{a} &=& \frac{1}{\max(A)} \label{eq:scale_s21} \\
c_{l} &=& c_{d} / c_{a} \label{eq:scale_s22}.
\end{eqnarray}
That is, we set the scale of diffuse albedo $c_a$ so that the largest albedo in the image is 1 and compute the coefficient of the lighting accordingly. To decide which strategy to use to compute the scale of lighting and albedo, we compute the following determinant when we regress $c_{d}$ and $c_{s}$: 
\begin{equation}
\mathcal{D} = \frac{(\tilde{I}_{d}\cdot \tilde{I}_{d}) (\tilde{I}_s \cdot \tilde{I}_s ) - (\tilde{I}_d\cdot \tilde{I}_s)^2 }{K} ,
\end{equation}
where $K$ is the number of pixels in the image. If $\mathcal{D} > 1e^{-7}$, we use \eqref{eq:scale_s11} and \eqref{eq:scale_s12}, otherwise we use \eqref{eq:scale_s21} and \eqref{eq:scale_s22} to compute the coefficient.

\vspace{-0.4cm}
\paragraph{Object insertion}
To render a new object into the scene, we first crop a planar region and pick a point on that plane to place the new object. The orientation, diffuse albedo and roughness value of the plane are all obtained from our predictions. We then render the plane and the object together using the lighting predicted at the point where we place the object. We render the plane and the new object together to ensure inter-reflections between them are properly simulated. We compute a high resolution environment map ($512\times 1024$) from the estimated spherical Gaussian parameters so that even very glossy material can be correctly handled. 

We render two images, $I_{\text{all} }$ and $I_{\text{pl}}$ and two binary masks, $M_{\text{obj}}$ and $M_{\text{all}}$. $I_{\text{all}}$ is the rendered image of plane and object and $I_{\text{pl}}$ is the rendered image of the plane only. $M_{obj}$ is the mask of the object and $M_{all}$ is the mask covered both the cropped plane and the object. We then edit the region of object and the region of cropped plane separately. Let $I$ be the original image and $I_{\text{new}}$ be the new image with the new rendered object. 
For the object region, we directly use the intensities as rendered in $I_{\text{all}}$ on the virtual object: 
\begin{equation}
I_{\text{new}} \odot M_{\text{obj}} = I_{\text{all}} \odot M_{\text{obj}}.
\end{equation}
For the remaining region on the plane, we blend in the original image intensities with the ratio of $I_{\text{all} }$ and $I_{\text{pl}}$:
\begin{equation}
I_{\text{new}} \odot (M_{\text{all} } - M_{\text{obj} }) = I \odot \frac{I_{\text{all}}}{I_{\text{pl}} } \odot (M_{\text{all} } - M_{\text{obj} }).
\end{equation}
All operations in the above relation are pixel-wise. This compositing procedure utilizes the idea of ratio (or quotient images) that has been used in the past for relighting~\cite{marschner1997inverse,shashua2001quotient}. It ensures that global effects due to object-plane interaction, such as soft shadows, are visualized (since they are rendered in $I_{\text{all} }$ but absent in $I_{\text{pl}}$), while keeping intensities consistent with the overall image. This suppresses high frequency artifacts in $I_{\text{all} }$ that might be caused by minor errors in estimation of albedo, roughness and lighting, thereby achieving greater photorealism.

Figures \ref{fig:object_insertion}, \ref{fig:early_demo_objInsert} and  \ref{fig:teaser} show several examples of object insertion on real images. In all these examples, we render white glossy objects with diffuse albedo $(0.8, 0.8, 0.8)$ and roughness value $0.2$. We use white color so that the color of lighting and global illumination effects can be clearly observed. We keep the shape simple and the roughness value low to better demonstrate the high frequency component in the predicted lighting. To better demonstrate our performance, a video of an object moving around the scene rendered by our prediction can be found at this \href{https://drive.google.com/file/d/1qD3xhK-NQuNu3ZbWeYE5annryZ9DCF3k/view?usp=sharing}{link}.

\begin{figure}[ht]
\centering
\includegraphics[width=1.0\columnwidth]{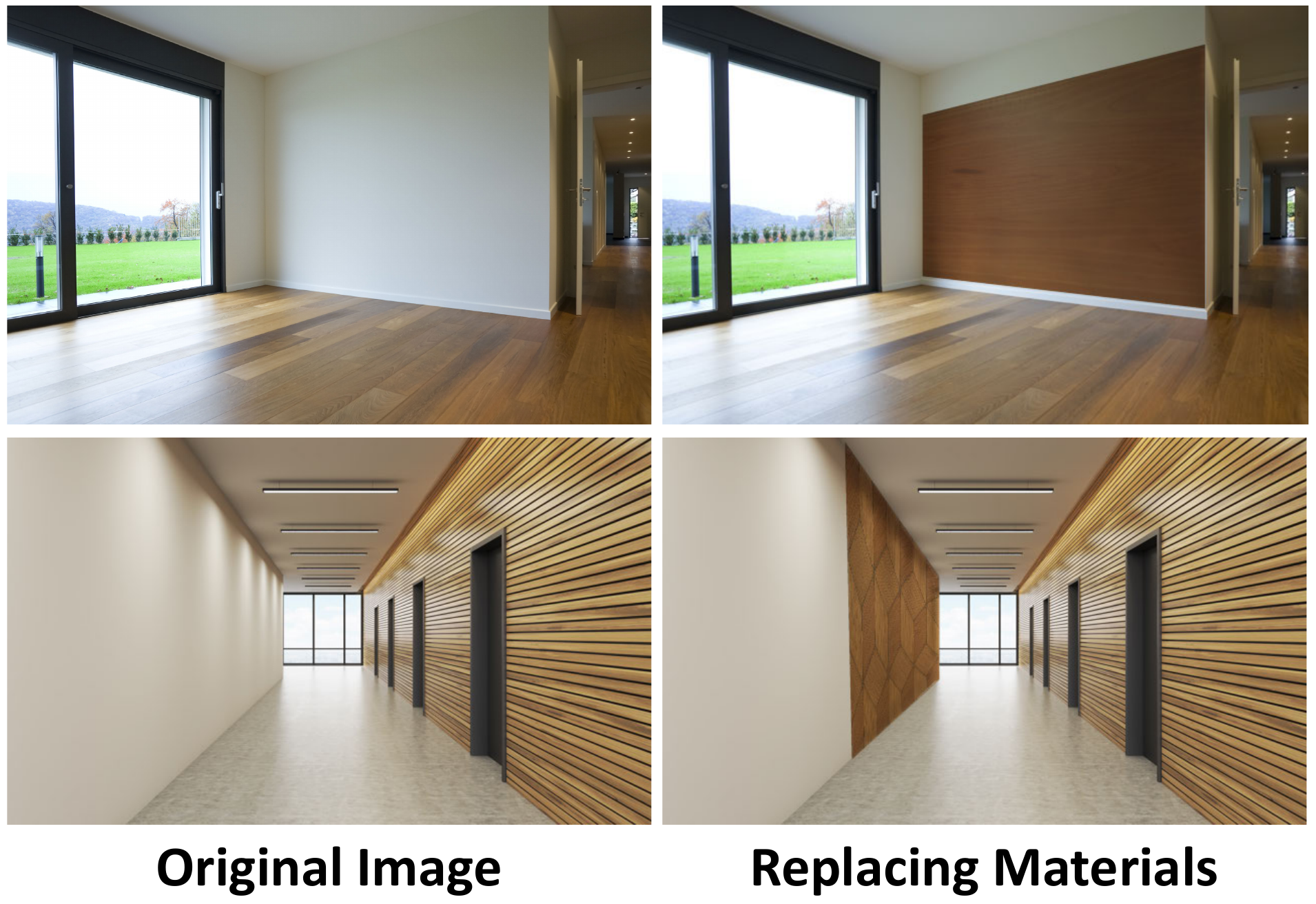}
\caption{\small Material editing. The left image is the original image and the right image is the rendered one with the material replaced in a part of the scene. We observe that the edited material looks photorealistic and even high frequency details from specular highlights and spatially-varying lighting are rendered well.}
\label{fig:replace_material}
\end{figure}

\begin{figure}[t]
\centering
\includegraphics[width=1.0\columnwidth]{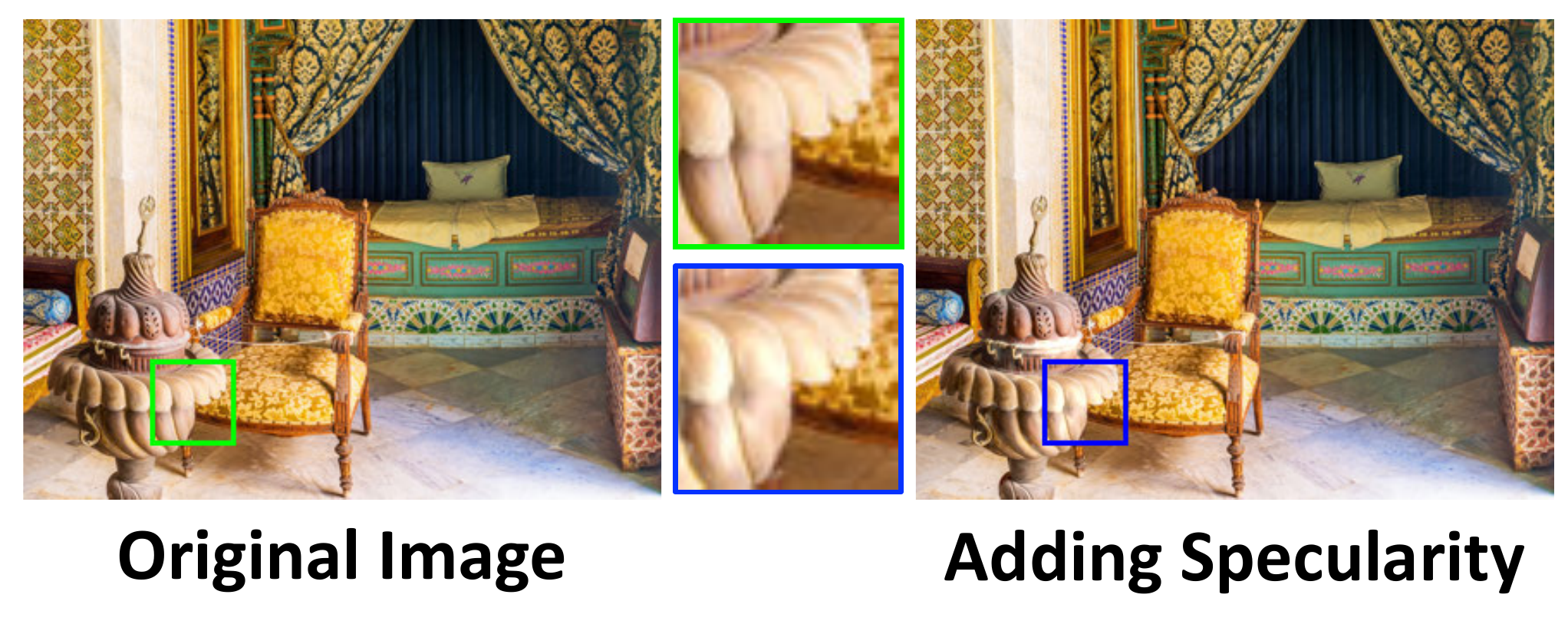}
\caption{\small Changing the specularity of an object. We keep the predicted geometry, diffuse albedo and spatially varying lighting as predicted, but change the roughness value to 0.2 and re-render the object, leading to more prominent specular highlights. }
\label{fig:add_specularity}
\end{figure}

\vspace{-0.4cm}
\paragraph{Material Editing}
Editing material properties of objects in a scene using a single photograph has applications for interior design and visualization. Our disentangled shape, material and lighting estimation allows rendering new appearances by replacing material and rendering using the estimated lighting. In Figure \ref{fig:early_demo_matEdit} and \ref{fig:replace_material}, we replace the material of a planar region with another kind of material and render the image using the predicted geometry and spatially varying lighting, where the spatially varying properties of the predicted lighting can be clearly observed. In the first example in Figure \ref{fig:early_demo_matEdit}, we can see the specular high light in the original image is preserved after changing the material, such specular high light effect can not be modeled by traditional intrinsic decomposition method since the direction of the incoming lighting is unknown. In Figure \ref{fig:add_specularity}, we add specular highlight to the selected object by changing the roughness value to 0.2 and render the object with predicted diffuse albedo, geometry and spatially varying lighting. We compute the residual image before and after changing the roughness value and add it back to the original image. Even though the difference is quite subtle, we observe that the distribution of the specular highlight looks plausible.

\section{Conclusion}
\label{sec::discussion}
We have presented the first holistic inverse rendering framework that estimates disentangled shape, SVBRDF and spatially-varying lighting, from a single image of an indoor scene. Insights from computer vision, graphics and deep convolutional networks are utilized to solve this challenging ill-posed problem. A GPU-accelerated renderer is used to synthesize a large-scale, realistic dataset with complex materials and global illumination. Our per-pixel SVSG lighting representation captures high frequency effects. Our network design imbibes intuitions such as a differentiable rendering layer, which are crucial for generalization to real images. Design choices such as a cascade structure and a bilateral solver lead to further benefits. Despite solving the joint problem, we obtain competitive results with respect to prior works that focus on constituent sub-problems, which highlights the impact of our dataset, representation choices and network design. We demonstrate object insertion and material editing applications on real images that capture global illumination effects, motivating applications in augmented reality and interior design.

\appendix

\section{Appendix Outline}
We have presented a method to automatically disentangle a single image of an indoor scene into its constituent physical scene factors -- geometry, spatially-varying reflectance, and illumination. In these appendices, we present more results, analyses and details. This includes: more challenging cases for our model (Appendix \ref{sec::Generalization} and Appendix \ref{sec:failureCases}), details about our SVBRDF model (Appendix~\ref{sec:BRDFModel}), dataset creation (Appendix \ref{sec:textureSynthesis}), our lighting model (Appendix~\ref{sec:ground-truthSphericalGaussianLobes}) and our network architecture and training details (Appendix~\ref{sec:trainingDetails}).

\section{Generalization to Outdoor Scenes}
\label{sec::Generalization}
\begin{figure*}
\centering
\includegraphics[width=1.0\linewidth]{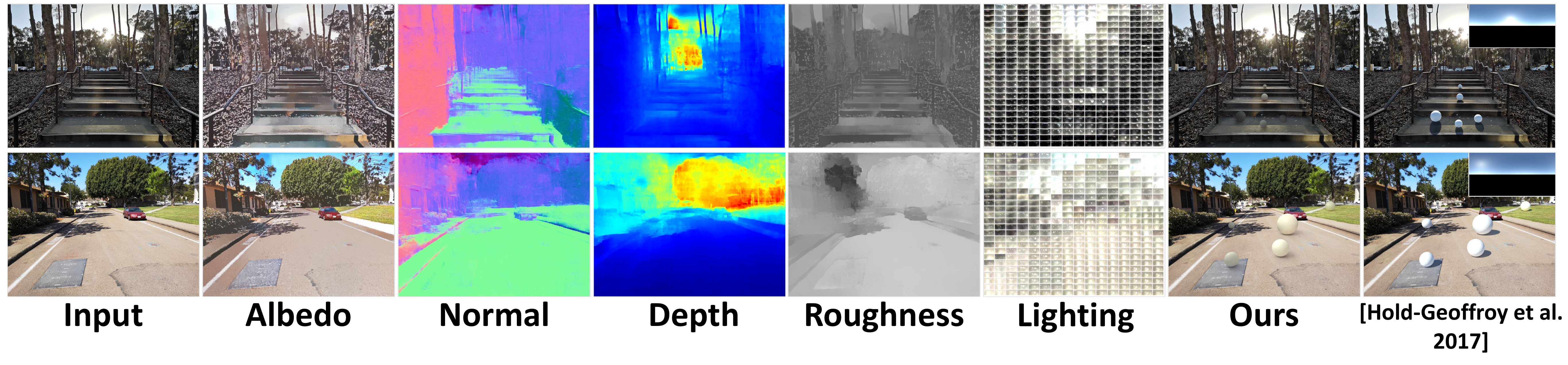}
\vspace{-0.1cm}
\caption{Our proposed method on outdoor scene. Even though our method is trained with synthetic indoor scenes only, it generalizes reasonably well to the outdoor scenes, which allows us to achieve reasonable object insertion results (the second last column) compared to the state-of-the-art \cite{hold-geoffroy2017outdoor} (the last column).   }
\label{fig:outdoor}
\end{figure*}

In this section, we test how well our model, which is trained with synthetic indoor scenes only, generalizes to outdoor scenes. The qualitative results are shown in Figure \ref{fig:outdoor}. While the network tries to interpret the outdoor scene into a room surrounded by walls, we observe that the overall estimation of geometry, lighting and diffuse albedo look reasonable. We also try to insert a new object into the scene using our predictions following the pipeline proposed in Section \ref{sec::experiments}, then compare with a state-of-the-art outdoor lighting estimation method \cite{hold-geoffroy2017outdoor}. As shown in the last two columns in Figure \ref{fig:outdoor}, the method of \cite{hold-geoffroy2017outdoor} can better preserve  high-frequencies in outdoor illumination, which results in shadows with hard boundaries, while our method tends to predict more low frequency lighting. This is probably due to the domain gap between training and test images. However, we notice that our model can usually predict the direction of the incoming light correctly and the spatial variation in the lighting prediction of our method looks much more realistic compared to \cite{hold-geoffroy2017outdoor}, which predicts a single lighting model for the whole image. 
\section{A Failure Case}
\label{sec:failureCases}
\begin{figure}[!!t]
    \centering
    \includegraphics[width=3.3in]{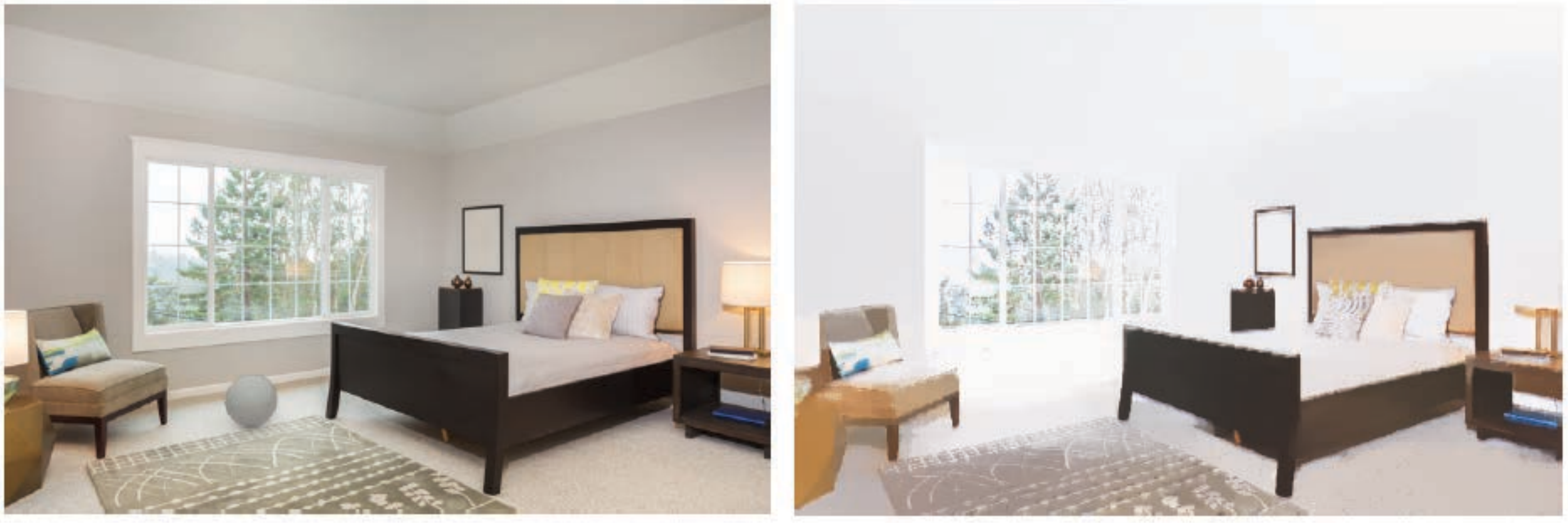}
    \caption{A failure case. (Left) The inserted object is rendered with an appearance that is darker than expected. (Right) The likely cause is an over-bright albedo estimation, which is traded off by a lighting estimate that has lower intensities.}
    \label{fig:ambiguity}
    \vspace{-0.1cm}
\end{figure}

While we observe largely successful object insertions in most experiments, some failure cases do occur. The ambiguity between albedo and lighting is a hard one to disentangle. In some cases, the albedo is estimated to be too bright (dark), with the lighting correspondingly estimated as too dark (bright). An example is shown in Figure \ref{fig:ambiguity}. Regardless, we emphasize that being able to estimate spatially-varying lighting along with SVBRDF and shape is an extremely hard problem, for which our network succeeds in an overwhelming number of experiments.

\section{BRDF Model and Material Categories}
\label{sec:BRDFModel}

\begin{figure}
\centering
\includegraphics[width=3.2in]{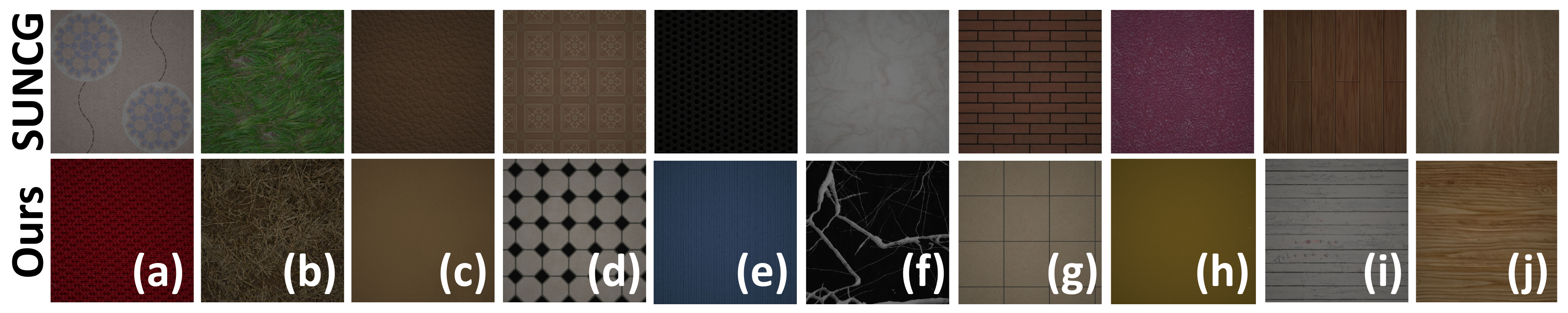}
\caption{The ten material categories and the corresponding spatially varying diffuse textures from both SUNCG dataset and our dataset. From left to the right: (a) fabric, (b) ground, (c) leather, (d) stone floor, (e) plastic, (f) stone specular, (g) stone wall, (h) wall paint, (i) wood floor, (j) wood. }
\label{fig:mat_category}. 
\end{figure}

\begin{table}
\begin{tabular}{|c|c|c|c|c|}
\hline
\small{wall paint} & \small{stone wall} & \small{leather} & \small{stone floor} & \small{plastic} \\
\hline
 127 &185  &10 &172 &94 \\
\hline
\small{stone specular} & \small{ground} & \small{fabric} & \small{wood floor} & \small{wood} \\
\hline
25 &243 &180 & 25 & 42 \\
\hline
\end{tabular}
\caption{The distribution of materials in our dataset, for the chosen semantic categories.}
\label{tab:mat_distribution}
\end{table}

\vspace{-0.1cm}
\paragraph{Our microfacet model}
We use a physically motivated microfacet BRDF model in our dataset. Let $A$, $N$ and $R$ be the spatially-varying diffuse albedo, normal and roughness, respectively. The BRDF model $f(l, v; A, N, R)$ is: 
\begin{eqnarray}
f(l, v; A, N, R)\!\!\!\!&=&\!\!\!\!f_d(l, v; A, N) + f_s(l, v; N, R) ,\\ 
f_{d}(v, l; A, N)\!\!\!\!&=&\!\!\!\!\frac{A}{\pi} , \\
f_{s}(v, l; N, R)\!\!\!\!&=&\!\!\!\! \frac{D(h, R)F(v, h)G(l, v, N, R)}{4(N\cdot l)(N\cdot v)} ,
\end{eqnarray}
where $f_d(\cdot)$ and $f_{s}(\cdot)$ are the diffuse and specular BRDF components. Here, $v$ and $l$ are view and lighting directions, and $h$ is the half angle vector, while $D(h, R)$, $F(v, H)$ and $G(l, v, h, R)$ are the distribution, Fresnel and geometric terms respectively, defined as
\begin{eqnarray}
D(h, R) &=& \frac{\alpha^{2}}{\pi\left[(N\cdot h)^{2}(\alpha^{2} - 1) + 1\right]^{2} } \nonumber  \\
\alpha &=& R^{2} , \nonumber \\
F(v, h) &=& (1 - F_{0} ) 2^{ -\left[5.55473(v\cdot h) + 6.8316\right](v\cdot h)} , \nonumber \\
G(l, v, R, N) &=& G_{1}(v, R, N)G_{1}(l, R, N) , \nonumber \\ 
G_{1}(l, R, N) &=& \frac{N\cdot l}{(N\cdot l)(1 -k) + k} , \nonumber \\
G_{1}(v, R, N) &=& \frac{N\cdot v}{(N\cdot v)(1 -k) + k} , \nonumber \\
k &=& \frac{(R + 1)^2}{8} . \nonumber 
\end{eqnarray}
We set $F_{0} = 0.05$ as suggested in \cite{karis2013unreal}.

\vspace{-0.1cm}
\paragraph{Material categories}
For mapping our materials on the SUNCG geometry in a manner consistent with semantics such as objects in the scene, we manually classified our dataset as well as the SUNCG materials into 10 categories. Samples from each dataset for these categories are shown in Figure \ref{fig:mat_category}. The number of material samples for each in our dataset are shown in Table \ref{tab:mat_distribution}.

\section{Tileable Texture Synthesis}
\label{sec:textureSynthesis}

\begin{figure}
\centering
\includegraphics[width=3.2in]{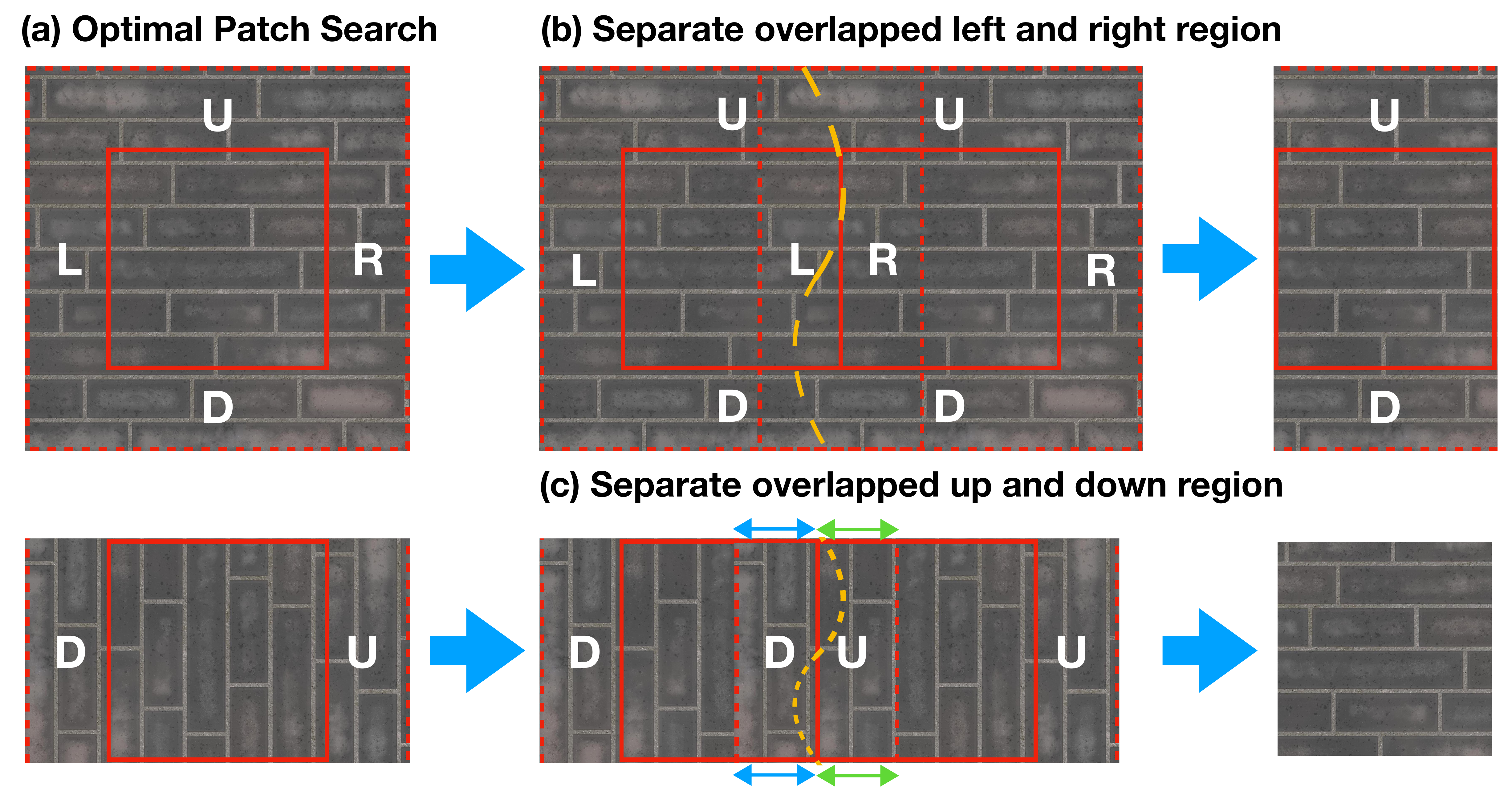}
\caption{The pipeline of using graph-cut based method for tileable texture synthesis. We first stitch the left-right boundaries of textures and then the up-down boundaries. When stitching up-down boundaries, we add a hard constraint so that left-right boundaries will remain tileable. } 
\label{fig:tileable_texture_pipeline}
\end{figure}

\begin{figure}
\centering
\includegraphics[width = 3.2in]{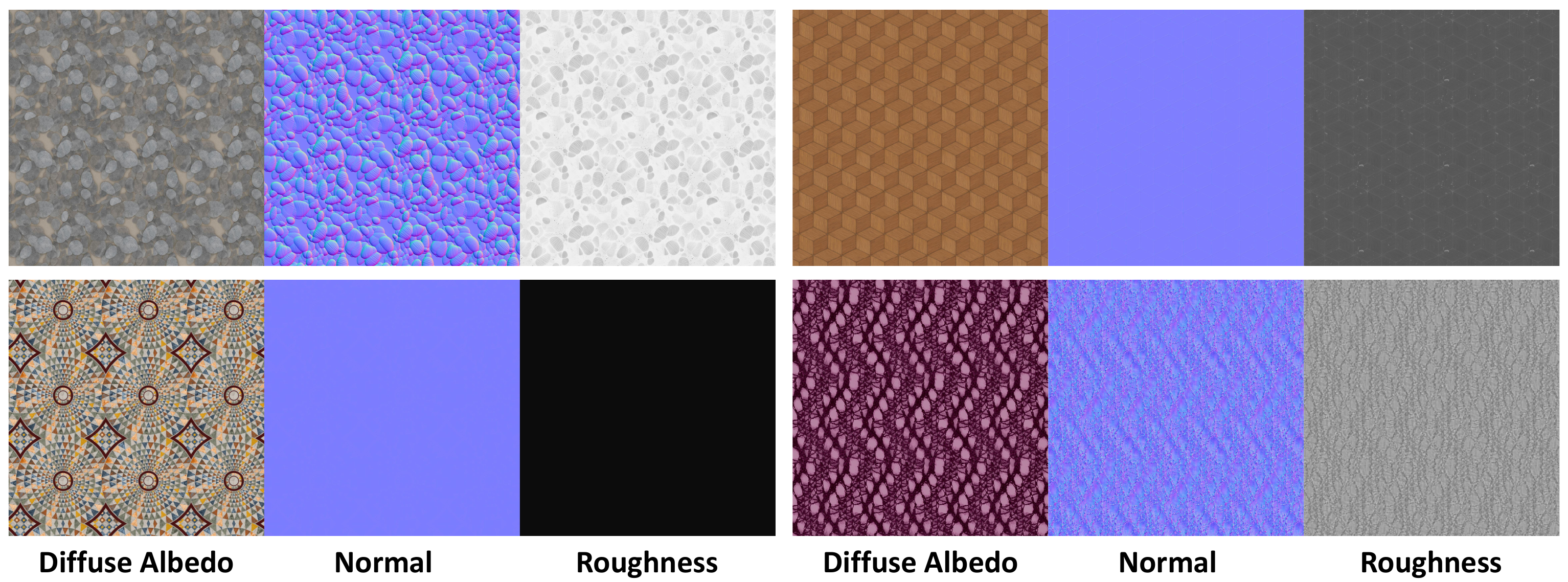}
\caption{Results of tileable texture synthesis. Each image is generated by tiling $3\times 3$ patches together. }
\label{fig:tiling}
\end{figure}
We use graph-cut based approach to generate tileable texture, which has the advantages of keeping the original texture structures \cite{kwatra2003graphcut}. The overall process is summarized in Figure \ref{fig:tileable_texture_pipeline}. We first crop smaller patch from the original SVBRDF textures and synthesize the cropped patch into tileable texture. Given the required size of the patch, we first globally search for the optimal patch by minimizing the gradient perpendicular to the boundary of the patches. More specifically, let $\mathcal{B}_x$
and $\mathcal{B}_{y}$ be the set of pixels on the horizontal and vertical boundaries of the patch. $A_{i}$, $N_{i}$ and $R_{i}$ are the diffuse color, normal and roughness at pixel $i$ respectively. We search for a patch so that
\begin{eqnarray}
&&\sum_{i\in \mathcal{B}_{y}}
\lambda_{A}\nabla_{x}A_{i} + \lambda_{N}\nabla_{x}N_{i} + 
\lambda_{R}\nabla_{x}R_{i} \nonumber\\
&+&\sum_{j\in \mathcal{B}_{x}}
\lambda_{A}\nabla_{y}A_{j} + \lambda_{N}\nabla_{y}N_{j} + 
\lambda_{R}\nabla_{y}R_{j}
\label{eq:crop_patch}
\end{eqnarray}
is minimized. The equation \eqref{eq:crop_patch} can be efficiently computed using integral graph. The overall complexity of finding optimal patch will be $\mathcal{O}(K)$ where $K$ is the number of pixels in the image. By minimizing \eqref{eq:crop_patch}, we avoid strong gradient near the boundaries of the patch, so that we can reduce artifacts in tileable texture synthesis. 

Once we find the patch, we crop not only the patch but also its surrounding regions. To make the patch tileable in $x$ direction, we overlap the right and left surrounding regions and use graph-cut method to find the best seam to separate the overlapping regions by minimizing a customized energy function. Unlike energy function in \cite{kwatra2003graphcut} which encourages the value of pixels at seam to be similar to the value of pixels in the original textures, our energy function encourages the gradients of pixels at the seam to be similar to the gradients of pixels in the original textures. As in \cite{kwatra2003graphcut}, we formulate the problem as a labeling problem. Let $I_{r, c}$ be an pixel in the overlapped texture map and $L_{r,c}\in \{1, 2\}$ be its label. With some abuse of notation, we define $I^{i}_{r, c}$ to be the value of pixel from patch $i,~i\in\{1, 2\}$. The gradient across the patches $i$ and $j$ is defined as $\nabla_{x} I^{i,j}_{r,c} = I^{j}_{r,c+1} - I^{i}_{r,c}$. Then the loss $\mathcal{L}_{I}(L_{r,c}=1, L_{r,c+1}=2)$ is defined as 
\begin{equation}
\min\left(\frac{||\nabla_{x}I^{1,2}_{r, c} - \nabla_{x}I^{1,1}_{r, c}||_{1}}{\max\left(||\nabla_{x}I^{1,1}_{r, c}||_{1}, 0.1\right)}, \frac{||\nabla_{x}I^{1,2}_{r, c} - \nabla_{x}I^{2,2}_{r, c}||_{1}}{\max\left(||\nabla_{x}I^{2,2}_{r, c}||_{1}, 0.1\right)}\right) \nonumber
\end{equation}
The final loss for is a weighted combination of losses from different texture map.
\begin{equation}
\lambda_{A}\mathcal{L}_{A} + \lambda_{N}\mathcal{L}_{N} + \lambda_{R}\mathcal{L}_{R}
\end{equation}
To make the texture tileable in $y$ direction, we repeat the above process by overlapping the up and down surrounding regions and finding the seam to separate them using graph-cut again. Notice that when separating the overlapping up and down regions, we need to make sure that the pixels at the right and left boundaries of the patches are from the same region so that the patch will remain tileable in x direction. We achieve this by adding an infinite smoothness term between every pair of pixels at the left and right boundaries in the same row so that they will always come from the same region. Figure \ref{fig:tiling} shows some texture synthesis examples. Each example is generated by tiling $3\times 3$ original patches together.  For each material from our dataset, we crop three patches of different sizes and the three patches will be considered as different materials in the following mapping SVBRDFs stage.
\section{Ground Truth Spherical Gaussian Lobes}
\label{sec:ground-truthSphericalGaussianLobes}

\begin{figure}
\centering
\includegraphics[width=3.2in]{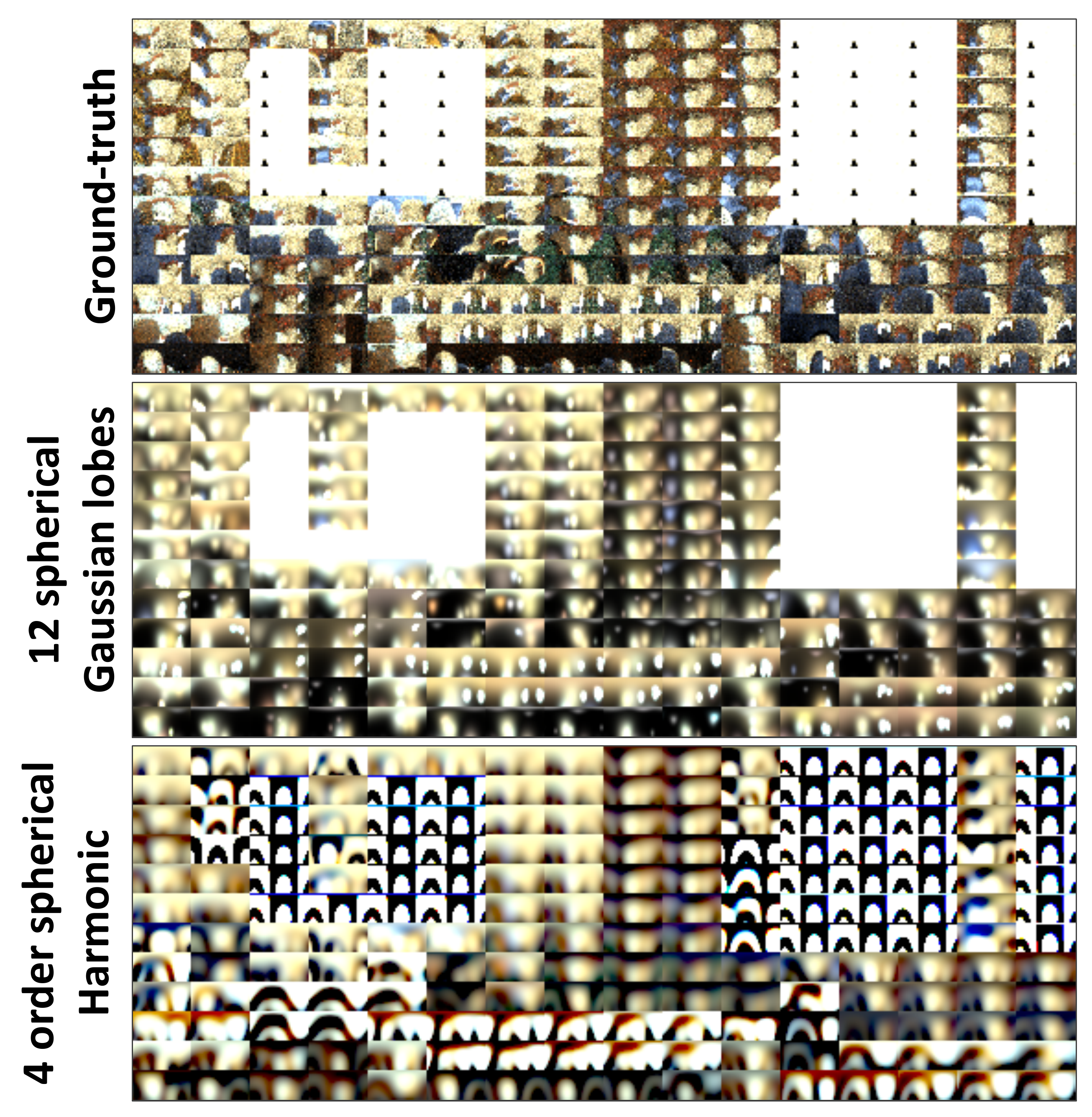}
\caption{Comparison of approximating lighting with spherical harmonics and spherical Gaussian. }
\label{fig:SGvsSH}
\end{figure}

We compute ground-truth spherical Gaussian lobe parameters by approximating the environmental lighting using the LBFGS method. These parameters are used to supervise spatially varying lighting prediction. We use 12 lobes to approximate per pixel lighting. To facilitate the training process, we assign an order to the 12 lobes by constraining each lobe to be in the certain range of the hemisphere. We roughly divide the hemisphere into $2\times 6$ regions. Following the notation in Section \ref{sec::networkDesign}, we define $\{\theta_{k}\}$, $\{\phi_{k}\}$, $\{\lambda_{k}\}$ and $\{F_{k}\}$ to be the spherical Gaussian parameters where 
\begin{equation}
\xi_{k} = (\sin\theta_{k}\cos\phi_{k}, \sin\theta_{k}\sin\phi_{k}, \cos\theta_{k}).
\end{equation}
In order to add the constraints, we reparameterize the spherical Gaussian parameters with $\{\hat{\theta}_{k}\}$, $\{\hat{\phi}_{k}\}$, $\{\hat{\lambda}_{k}\}$ and $\{\hat{F}_{k}\}$ such that
\begin{eqnarray}
\lambda_{k} &=& \exp{(\hat{\lambda}_{k}) } ,\\
F_{k} &=& \exp{(\hat{F}_{k})} , \\
\theta_{k} &=& a \tanh(\hat{\theta}_{k}) + b_{k} , \\
\phi_{k} &=& c \tanh(\hat{\phi}_{k}) + d_{k} ,
\end{eqnarray}
where $a = \frac{3\pi}{8}$ and $c = \frac{\pi}{2}$ are scaling factors. Here, $b_{k}$ and $d_{k}$ are offset parameters that are computed as 
\begin{eqnarray}
b_k &=& \frac{\pi}{4}(k \bmod 2 + \frac{1}{2}) , \\
d_k &=& \frac{\pi}{3}(k \bmod 6 + \frac{1}{2}) - \pi .
\end{eqnarray}
The initialization of the parameters are $\hat{\theta}_{k} = 0$, $\hat{\phi}_{k} = 0$, $\hat{F}_{k} = 0$ and $\hat{\lambda}_{k} = \log(\frac{\pi}{2})$. The loss function is the log-encoded loss as described in \eqref{eq:lossDepth}. 

Figure \ref{fig:SGvsSH} compares using spherical Gaussian and spherical harmonics to approximate the spatially varying lighting, which corresponds Figure \ref{fig:lightApprox} and Table \ref{tab:lightApprox} in Section \ref{sec::networkDesign}. It is clearly observed that with a similar number of parameters, spherical Gaussians can better recover high frequency effects, resulting in a reconstructed spatially-varying lighting closer to ground truth. 
\section{Network Structures and Training Details}
\label{sec:trainingDetails}

\begin{figure*}
\centering
\includegraphics[width=6.4in]{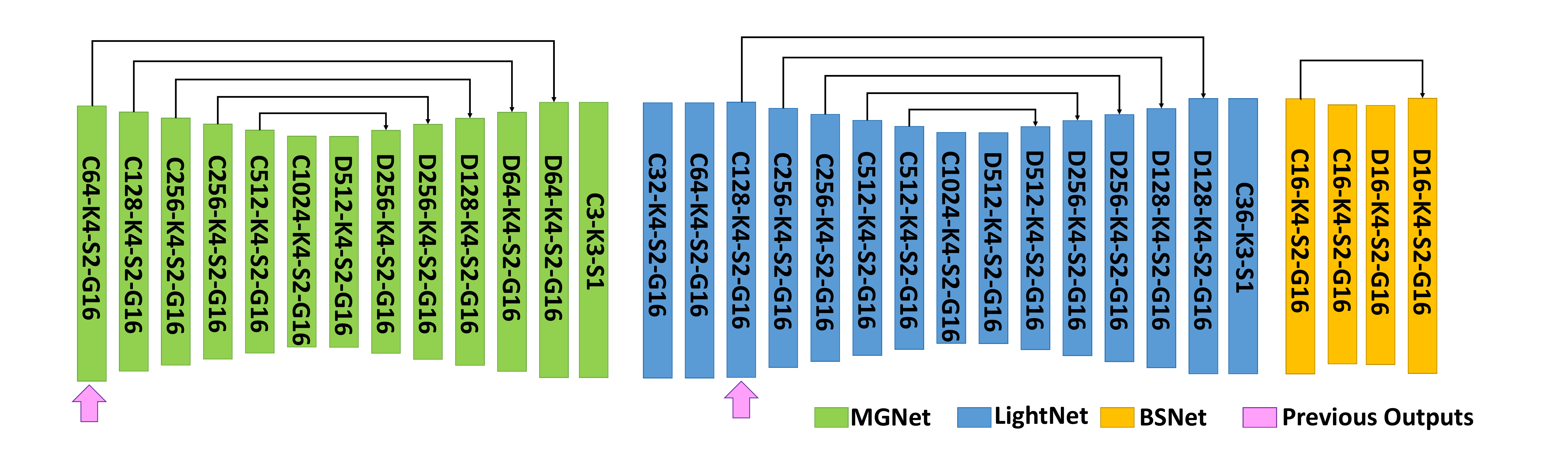}
\caption{Our network architectures. $\mathbf{C/D}X$-$\mathbf{K}Y$-$\mathbf{S}Z$-$\mathbf{G}W$ represents a convolution or transpose convolution layer with channel number $X$, kernel size $Y$, stride $Z$ and group normalization with $W$ channels in a group. 
$\mathbf{Previous~Outputs}$ represents the place where we concatenate the predictions of previous level of cascade with current features. }
\label{fig:networkArc}
\end{figure*}

\renewcommand{\arraystretch}{1.2}
\begin{table*}[t]
\centering
\begin{tabular}{|l|c|c|c|c|c|c|c|c|c|c|c|c|c|c|}
\hline
& $\alpha_{A}$ & $\alpha_{N}$ & $\alpha_{R}$ & $\alpha_{D}$ & $\alpha_{L}$ & $\alpha_{ren}$ & $\alpha_{\lambda}$ & $\alpha_{\xi}$ & $\alpha_{F}$ & epochs & iters. & $\text{lr}_{\mathbf{MG}}$ & $\text{lr}_{\mathbf{Light}}$ & batch  \\
\hline
$\mathbf{MGNet}_0$ & 1.5 & 1.0 & 0.5 & 0.5 & - & - & - & - & - & 21 & - & $1e^{-4}$ & - & 16\\
\hline
$\mathbf{LightNet}_0$ & - & - & - & - & 10 & 10 & $5e^{-4}$ & 1 & 0.5 & 15 & - & - & $1e^{-4}$ & 4  \\
\hline
$\text{Fine Tune}_0$ & 7.5 & 5.0 & 2.5 & 2.5 & 10 & 10 & $5e^{-4}$ & 1 & 0.5 & - & 4000 & $1e^{-9}$ & $1e^{-6}$ & 4 \\
\hline
$\mathbf{MGNet}_1$ & 1.5 & 1.0 & 0.5 & 0.5 & - & - & - & - & - & 8 & - & $1e^{-4}$ & - & 6\\
\hline
$\mathbf{LightNet}_1$ & - & - & - & - & 10 & 10 & $5e^{-4}$ & 1 & 0.5 & 8 & - & - & $1e^{-4}$ & 4\\
\hline
$\text{Fine Tune}_1$ & 7.5 & 5.0 & 2.5 & 2.5 & 10 & 10 & $5e^{-4}$ & 1 & 0.5 & - & 4000 & $1e^{-9}$ & $1e^{-6}$ & 3   \\
\hline
\end{tabular}
\vspace{0.1cm}
\caption{Hyper parameters for training $\mathbf{MGNet}_{i}$ and $\mathbf{LightNet}_{i}$.  }
\label{tab:training_details}
\end{table*}

\begin{table}[]
\centering
\begin{tabular}{|c|c|c|c|c|c|c|}
\hline
& $\alpha_{A}$ & $\alpha_{R}$ & $\alpha_{D}$ & iters. &  $\text{lr}_{\mathbf{BS}}$ & batch \\
\hline
$\mathbf{BSNet}$ & 1.5 & 0.5 & 0.5 & 600 & $1e^{-4}$ & 6 \\
\hline
\end{tabular}
\vspace{0.1cm}
\caption{Hyper parameters for training $\mathbf{BSNet}$. }
\label{tab:training_details_BS}
\end{table}

The network structures are summarized in Figure \ref{fig:networkArc}. Note that we use group normalization \cite{wu2018group} instead of batch normalization so that we can train the network with smaller batch size. The padding size is dynamically assigned according to the feature map size so that the feature maps after up-sampling can be aligned with the feature maps coming from skip links. Therefore, our network can process image of arbitrary size without scaling and cropping. The network for spatially varying lighting predictions has more parameters because we find it necessary to achieve reasonable performances for this task. 

We use Adam optimizer to train our network. Each level of cascade network is trained separately. To train cascade network of level $i$, we first train $\mathbf{MGNet}_{i}$ and $\mathbf{LightNet}_{i}$ separately and then fine-tune them together. The loss function to train $\mathbf{MGNet}_{i}$ is 
\begin{equation}
\alpha_{A}\mathcal{L}_{A} + \alpha_{N}\mathcal{L}_{N} + \alpha_{R}\mathcal{L}_{R} + \alpha_{D}\mathcal{L}_{D},
\end{equation}
with various terms as defined in Sec. \ref{sec::networkDesign}. We add the rendering loss $\mathcal{L}_{ren}$ when training $\mathbf{LightNet}_{i}$. The loss function to train $\mathbf{LightNet}_{i}$ is 
\begin{equation}
\alpha_{L}\mathcal{L}_{L} + \alpha_{ren}\mathcal{L}_{ren} + \sum_{k=1}^{K}\alpha_{\lambda}\mathcal{L}_{\lambda_{k}} + \alpha_{\xi}\mathcal{L}_{\xi_{k}} + \alpha_{F}\mathcal{L}_{F_{k}} .
\end{equation}
The loss function for fine-tuning the whole pipeline is defined in Eq. \eqref{eq:loss}. Finally, the loss function to train $\mathbf{BSNet}$ is 
\begin{equation}
\alpha_{A}\mathcal{L}_{A} + \alpha_{R}\mathcal{L}_{R} + \alpha_{D}\mathcal{L}_{D}.
\end{equation}

All other hyper parameters including initial learning rate, training epochs and and coefficients $\alpha_{(\cdot)}$ are summarized in Table \ref{tab:training_details} and Table \ref{tab:training_details_BS}. The learning rates are decreased by half every 10 epochs. 

\vspace{-0.1cm}
\paragraph{Fine-tuning on real datasets}
We use similar strategy to fine-tune on IIW dataset \cite{bell2014intrinsic} and NYU dataset \cite{silberman2012indoor}. We take the trained model and fine-tune each level of cascade sequentially. 
The learning rate is $1e^{-4}$ and the batch size is 4 for the first level of cascade and 2 for the second level. In each iteration, we send two batches of images to the network, one from our synthetic dataset and the other from the real dataset. The $\mathbf{MGNet}_{i}$ and $\mathbf{LightNet}_{i}$ are trained in an end-to-end manner. The loss function for images from our dataset is the same as Eq. \eqref{eq:loss}. The loss function for fine-tuning on NYU dataset is just a combination of the loss on each component. We add the rendering loss by comparing the rendered image with the input image:
\begin{equation}
\alpha_{ren}\mathcal{L}_{ren} + \alpha_{N}\mathcal{L}_{N} + \alpha_{D}\mathcal{L}_{D}.
\end{equation}
When fine-tuning on IIW dataset, we include ordinal reflectance loss $\mathcal{L}_{ord}$ which is the same as defined in \cite{li2018cgintrinsics}. The loss function for images from IIW dataset is
\begin{equation}
\alpha_{ren}\mathcal{L}_{ren} + \alpha_{ord}\mathcal{L}_{ord}.
\end{equation}
When training on NYU dataset, we also do data augmentation by randomly flipping, cropping and scaling the input images with a scale uniformly sampled from 0.8 to 1.2 since the dataset is relatively small.

{\small
\bibliographystyle{ieee}
\bibliography{refs}
}

\end{document}